\documentclass[10pt,twocolumn,letterpaper]{article}

\usepackage[pagenumbers]{cvpr} 

\usepackage{microtype}
\usepackage{graphicx}
\usepackage{booktabs} 
\usepackage{multirow}
\usepackage{pifont}
\usepackage{colortbl}
\usepackage{xcolor}
\usepackage{placeins}
\usepackage{tabularx}
\usepackage{pifont}
\usepackage{bbding}
\usepackage{fontawesome}
\usepackage{mathtools}
\usepackage[table]{xcolor}

\definecolor{RLDshade}{RGB}{237,244,251}
\definecolor{BaselineGray}{RGB}{105,112,120}

\usepackage{amsmath,amsfonts}
\usepackage{algorithm}
\usepackage{algpseudocode}
\usepackage{array}
\usepackage{textcomp}
\usepackage{stfloats}
\usepackage{url}
\usepackage{verbatim}
\usepackage{tikz}
\usetikzlibrary{positioning}

\definecolor{cvprblue}{rgb}{0.21,0.49,0.74}
\usepackage[pagebackref,breaklinks,colorlinks,allcolors=cvprblue]{hyperref}

\def\confName{CVPR}
\def\confYear{2027}

\title{Rectifying Latent Dynamics inside Looped Transformers}

\author{
    Jizhihui Liu$^{1,2}$\quad Mingyue Guo$^1$\thanks{Corresponding author.}\quad Jinpeng Wang$^2$\footnotemark[1]\quad Lu Wang$^3$\quad Yaowei Wang$^{1,2}$\\
    $^1$PengCheng Laboratory \quad
    $^2$Harbin Institute of Technology, Shenzhen \\
    $^3$University of Chinese Academy of Sciences \\
    {\small \tt danielement321@gmail.com \quad guomy@pcl.ac.cn \quad  wangjp26@gmail.com}
}

\begin{document}
\maketitle

\begin{abstract}

Looped transformers iteratively refine token representations in the hidden space, providing a promising way for latent reasoning. However, the test-time scaling of looped transformers is limited by the overthinking issue: the performance first peaks and then drops with the growth of loop iterations. We analyse the latent dynamics and identify two potential causes for this phenomenon: 1) an excessively large update in the first loop directly leads tokens into the wrong space; 2) the persistent and stochastic late-stage drift pushes the tokens from right to wrong. To address these issues, we propose Rectifying Latent Dynamics (RLD), an explicit regulation approach comprising three components: \textbf{1) Recurrent Depth Modulator} to inject recurrence depth signal and dynamically control the update magnitude according to token difficulty and loop iteration; \textbf{2) Latent Residual Rectifier} to capture evolving update directions and rectify distorted traces, making the trace smoother and more coherent; \textbf{3) Update Clipping} to constrain and slow down excessive latent updates and underscore step-by-step reasoning potential. With fewer than 5M trainable parameters, RLD is easy to deploy and generalizes well to multiple tasks and unseen recurrence depths. We evaluate RLD on both language and vision-language models across 18 benchmarks. Specifically, the vanilla LoopVL model suffers an 80.7\% performance degradation on ChartQA when the inference depth is $2\times$ the default depth, while RLD preserves 95.2\% of the vanilla model score at default depth. For looped language models, standard SFT faces a 77.2\% drop on GSM8k when the depth exceeds the training depth, while RLD only loses 0.8\% of the highest score under this setting.

\end{abstract}

\section{Introduction}

\begin{figure}[t]
    \centering
    \includegraphics[width=\linewidth]{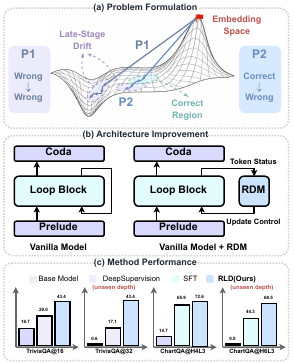}
    \caption{\textbf{(a) Problem Formulation:} Looped transformers tend to have a large update in the first loop iteration, and the subsequent loops reflect a persistent late-stage drift in the latent dynamics. \textbf{(b) Architecture Improvement:} We parallel an RDM module to provide an explicit depth signal to control the update magnitude. \textbf{(c) Method Performance:} The performance of Ouro and LoopVL, the number after @ denotes the inference-time loop iteration.}
    \label{fig:intro_performance_trajectory}
\end{figure}

Reasoning is one of the key pursuits of artificial intelligence. Explicit Chain-of-Thought reasoning \cite{wei2022chain, guo2025deepseek, kojima2022large, wang2022self} enables Large Language Models (LLMs) \cite{grattafiori2024llama, yang2025qwen3} to think in the space of natural language by generating abundant thinking tokens before the final answer. Intuitively, models do not necessarily need to think in the way we humans do. Looped transformers \cite{geiping2026scaling, qian2026loopvl, zhu2025scaling} extend the reasoning space from language to parameters by refining the tokens with a set of transformer blocks \cite{vaswani2017attention} time by time, providing a promising way to enable latent reasoning. However, the inference-time scalability of looped transformers faces an overthinking issue. As shown in Fig. \ref{fig:intro_performance_trajectory}(c), more loops do not necessarily result in better performance. Moreover, the optimal loop setting varies between tasks (Table \ref{tab:performance_comparison}), making it infeasible to find a gold value for all the tasks.

Since we cannot set a single value suitable for all the tasks, a natural idea is to stabilize the model performance after reaching the peak. STARS \cite{pmlr-v306-yang26t} trains the model with Jacobian spectral radius regularization to force the model to converge toward stable fixed points. However, this method modifies the already-learned knowledge in the model and faces performance degradation under unseen or challenging tasks. RL-Halting \cite{kuo2026stabilizing} and TaH \cite{pmlr-v306-fu26t} focus on dynamic depth and deciding when to stop for a token. These methods improve robustness to different computation budgets, while they do not necessarily ensure extrapolability to unseen recurrence depths. InfiLoop \cite{li2026scaling} and \cite{zhou2026learning} preserve intermediate states for deeper reference and achieve depth extrapolation. Nevertheless, they require training a model from scratch, and their efficacy on billion-scale looped models remains largely unexplored. To mitigate these issues, we investigate approaches that make modifications to the already-learned knowledge in the base model as little as possible and remain robust to unseen tasks and recurrence depths.

\begin{figure}[t]
    \centering
    \includegraphics[width=\linewidth]{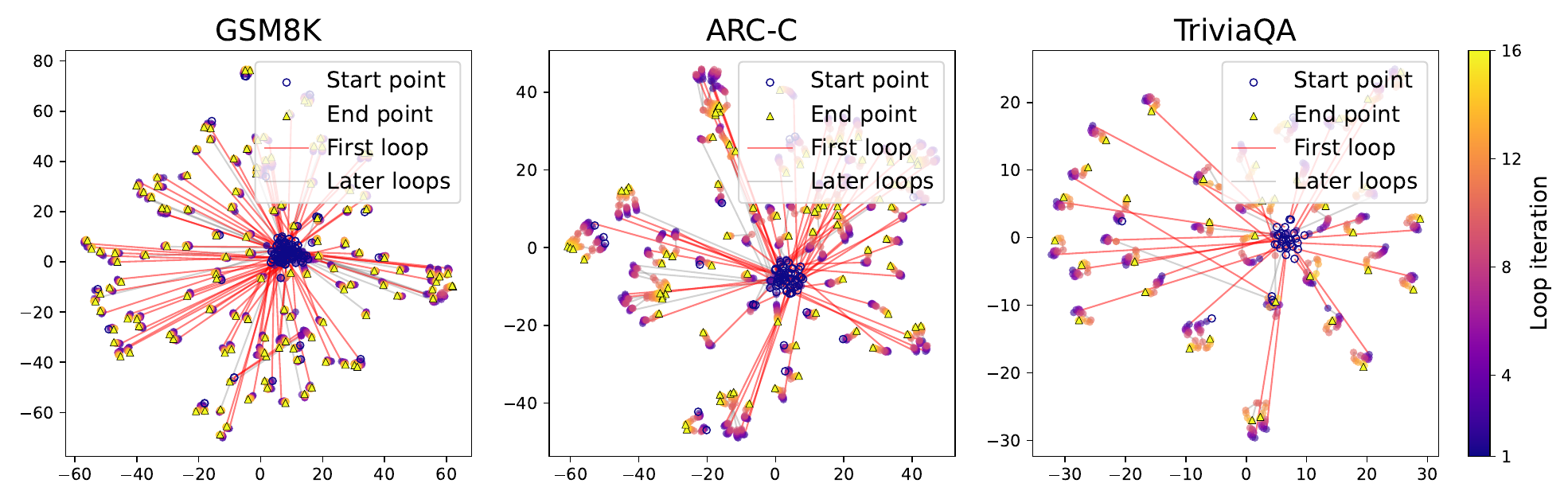}
    \caption{\textbf{Trajectory Visualization of Different Tasks.} We plot the trace of tokens from various tasks using t-SNE \cite{JMLR:v9:vandermaaten08a}. The number of loops is set to 16. More examples can be found in Appendix \ref{sec:supp_vis_task_traj}.}
    \label{fig:intro_tarj_vis}
\end{figure}

To understand the cause of overthinking in looped transformers, we first study the latent dynamics in the model. We randomly choose three representative samples: one math problem from GSM8k \cite{cobbe2021training}, one science reasoning problem from ARC-Challenge \cite{clark2018think}, and one factual knowledge retrieval problem from TriviaQA \cite{joshi2017triviaqa}, and plot the latent trajectory in Fig. \ref{fig:intro_tarj_vis}. Intuitively, difficult tasks require the model to think step by step and reason for multiple steps before approaching the final answer, while easier tasks need less thinking, and the model can get the final answer quickly in a few steps. However, the visualization reveals a phenomenon contradictory to that intuition: \textit{the model tends to have quick, one-shot thinking for harder tasks like GSM8k, as the first loop almost takes up all the update of one token; while for easier tasks like TriviaQA, the model always hesitates and keeps changing the answer time by time.} This observation provides two potential paths for a wrong token: \textbf{P1:} The model tends to finish thinking in one go, and once the first loop goes wrong, the subsequent loops can make little difference and have almost no chance to correct it. This limits the model's potential to think deeper with more loops. \textbf{P2:} Even if the first loop succeeds in taking the token to the correct region, the accumulated late-stage drift and hesitation may push the token to escape, finally leading to a wrong answer. We denote these two paths as ``wrong $\rightarrow$ wrong'' and ``correct $\rightarrow$ wrong'', and illustrate them in Fig. \ref{fig:intro_performance_trajectory}(a).

The key to solving the aforementioned issues lies in two aspects: \textbf{(1)} Enabling the model to have a direct perception of loop steps so that it can adjust the update magnitude according to recurrence depth. \textbf{(2)} Rectifying the update direction to avoid the late-stage drift in deeper loops. To this end, we propose a \textbf{Recurrent Depth Modulator} in parallel to the loop block to provide an explicit depth signal. The modulator updates a depth vector along with hidden states, and controls the update magnitude based on recurrence depth and token type. To rectify the late-stage drift in latent dynamics, we introduce a \textbf{Latent Residual Rectifier} to fix the direction of each update. Drawing inspiration from gradient clipping in optimizer design \cite{pmlr-v80-shazeer18a}, we adopt \textbf{Update Clipping} to constrain extremely large update steps and force the model to think ``slowly'' and step by step in the latent space. The only training parameters in RLD are two 3-layer MLPs and one linear projection, while the model is always frozen.

We evaluate RLD on both LLM and VLM, and show its stability under high loops and generalization ability across unseen depths and tasks. Specifically, the vanilla LoopVL model suffers from a 91.6\% and 80.7\% performance degradation on the AI2D and ChartQA respectively when the inference-time loop is $\times2$ the training depth, while RLD preserves 95.6\% and 95.2\% results on these settings. On depth settings unseen during training, standard SFT faces a 77.2\% drop on GSM8k, while RLD only loses 0.8\% of the highest score, showing outstanding extrapolation ability.

Our main contributions are summarized as follows:

\begin{itemize}
    \item We analyse the overthinking in looped models and propose an explanatory hypothesis from the latent dynamics.
    \item We design a \textbf{Recurrent Depth Modulator} and \textbf{Update Clipping} to control the update magnitude of latent dynamics, which helps stabilize performance after the peak and avoid shallow in-one-go thinking.
    \item We introduce a \textbf{Latent Residual Rectifier} to reduce the late-stage drift of tokens in later loop stages, making the trajectory smoother and more continuous.
    \item We evaluate RLD on both LLMs and VLMs, proving its ability to maintain performance after the peak and generalization to unseen depths or tasks.
\end{itemize}

\section{Related Works}
\paragraph{Looped Transformers.} Looped Transformers \cite{dehghani2019universal,lan2020albert,saunshi2025reasoning} decouple computational depth from model parameters by repeatedly applying shared Transformer blocks. They have achieved comparable or even superior performance to larger models across various tasks, demonstrating the effectiveness of this paradigm. While LLMs improve reasoning capabilities through explicit chain-of-thought (CoT), looped Transformers offer an alternative approach by performing iterative reasoning in latent space. Huginn \cite{geiping2026scaling} adopts a three-stage architecture consisting of a prelude, a recurrent block, and a coda. The recurrent block iteratively refines the hidden state conditioned on both the input representation and the current state. In contrast, Ouro \cite{zhu2025scaling} removes the input injection and refines the hidden state solely through recurrent computation. More recently, LoopVL \cite{qian2026loopvl} extends looped Transformers to multimodal settings, demonstrating their effectiveness in vision-language modeling. We evaluate our method on both Ouro and LoopVL, covering language-only and multimodal settings. Notably, our method is model-agnostic and is not restricted to these specific architectures.

\paragraph{Dynamic Computation and Depth Robustness.} Looped Transformers enable dynamic computation by adjusting the number of recurrent iterations for simple and challenging inputs. Ouro~\citep{zhu2025scaling} learns an exit gate to determine when to terminate recurrence, while Mixture-of-Recursions~\citep{bae2025mor} employs lightweight routers to assign different recursion depths to individual tokens. RL-Halting \citep{kuo2026stabilizing} learns a stochastic stopping policy during training to improve stability across different loop counts. Inspired by shortcut models in image generation, LoopFormer~\citep{LoopFormer} introduces a shortcut-consistency objective to align representations across different computation depths. Compared with the early-exit-based dynamic computation methods, research on depth robustness focuses on stabilizing the model performance across inference-time loop depths. Among them, STARS \cite{pmlr-v306-yang26t} adds a Jacobian spectral regularization during tuning to encourage convergence. DeepSupervision \cite{chng2026looped} supervises the output of both the final and internal hidden states. In this work, we introduce two lightweight modules and a clipping policy to schedule and stabilize model performance across varying depths without retraining the pretrained model.

\section{Method}
RLD is primarily composed of two modules: Recurrent Depth Modulator (\cref{sec:method_RDM}) and Latent Residual Rectifier (\cref{sec:method_LRR}), and one policy: Update Clipping (\cref{sec:method_UC}). The overall pipeline is shown in Algorithm \ref{alg:rld}.

\subsection{Recurrent Depth Modulator}
\label{sec:method_RDM}

\begin{figure}[t]
    \centering
    \includegraphics[width=\linewidth]{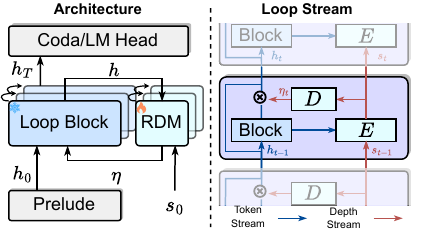}
    \caption{\textbf{Overview of Recurrent Depth Modulator.} RDM is an encoder-decoder module parallel to the original loop blocks. It maintains a depth status $s$ during the loop to provide an explicit control signal to the residual stream. During training, the base model is always frozen and only the additional module is updated.}
    \label{fig:method_RDM}
\end{figure}

Token updates in looped models are quite similar to parameter updates in optimizers \cite{robbins1951stochastic, DBLP:journals/corr/KingmaB14, loshchilov2017sgdr}; however, it remains fixed during all iterations, making it unable to adapt to different recurrent phases. To enable the model to perceive thinking depths and dynamic update control, we introduce the Recurrent Depth Modulator (RDM), as shown in Fig. \ref{fig:method_RDM}.

Two key design pursuits of RDM are: \textbf{(1) Extrapolation:} The model should work stably on unseen depths during training. \textbf{(2) Adaptivity:} Update magnitude differs between simple and hard tokens. It is notable that a recurrent architecture $\mathcal{F}$ already features extrapolation after stability: when the input reaches a stable point $h_{t^*}$ at timestep $t^*$, the output has $\mathcal{F}(h_{t^*})\approx h_{t^*}$, and this trend stabilizes at any subsequent timestep $t>t^*$ ideally. For adaptivity, the RDM must have a mutual interaction with the loop blocks to control tokens with varying difficulty, which is formulated in Eq. \ref{eq:method_interaction}.

\begin{equation}
    \text{RDM}
    \xrightleftharpoons[
        \text{Block Output}~\mathbf Z_t
    ]{
        \text{Update Control}~\boldsymbol\eta_t
    }
    \text{Model}.
    \label{eq:method_interaction}
\end{equation}

To this end, RDM is a module parallel to the loop block and updates depth states simultaneously with token states. Let $\mathbf{H}_t \in \mathbb{R}^{N\times d}$ and $\mathbf{S}_t \in \mathbb{R}^{N\times d_s}$ denote the token and depth states at loop $t$, where $N$, $d$, $d_s$ stand for token number, token dimension and depth dimension. RDM first fuses the two and updates the depth state with an encoder $\mathcal{E}^{\text{RDM}}$, and then decodes the updated depth states into a magnitude controller $\boldsymbol\eta_t \in \mathbb{R}^{N\times 1}$ with a decoder $\mathcal{D}^{\text{RDM}}$. The update of depth and token stream can be formulated by

\begin{equation}
    \mathbf{Z}_t=\mathcal{F}^{\text{L-blk.}}(\mathbf{H}_{t-1}),
\end{equation}
\begin{equation}
    \mathbf{S}_t=\mathcal{E}^{\text{RDM}}\left([\operatorname{LN}(\mathbf Z_t)\Vert \operatorname{LN}(\mathbf S_{t-1})]\right) + \mathbf{S}_{t-1},
\end{equation}
\begin{equation}
    \boldsymbol\eta_t=
    2\sigma\!\left(\mathcal D^\text{RDM}(\mathbf S_t)\right)
    \in(0,2)^{N\times1},
\end{equation}

\noindent where $\sigma$ is the sigmoid function, $\text{LN}$ denotes layer normalization \cite{DBLP:journals/corr/BaKH16}, and $\Vert$ is concatenation. This design ensures the update magnitude can be adjusted larger or smaller according to the depth, and considers token difficulty for dynamic scheduling. After obtaining $\boldsymbol{\eta_t}$, the update is modulated by

\begin{equation}
    \boldsymbol\Delta_t=\mathbf Z_t-\mathbf H_{t-1},
    \label{eq:method_original_update}
\end{equation}
\begin{equation}
    \mathbf H_t=
    \mathbf H_{t-1}
    +\boldsymbol\eta_t\odot\boldsymbol\Delta_t.
    \label{eq:method_RDM_original_update}
\end{equation}

\subsection{Latent Residual Rectifier}
\label{sec:method_LRR}

\begin{table}[t]
\centering
\caption{\textbf{Performance Comparison between Tuning Methods.} We follow the same training recipe as \cref{sec:exp_setup} except using 10\% training data. Different trainable parameters of LRR are achieved by adjusting the MLP size. The number denotes loop iterations.}
\label{tab:method_LRR}
\begin{tabular}{lccccc}
\toprule
\multirow{2}{*}{\textbf{Method}} &
\multirow{2}{*}{\textbf{Param.}} &
\multicolumn{2}{c}{\textbf{Winogrande$\uparrow$}} &
\multicolumn{2}{c}{\textbf{CoQA$\uparrow$}} \\
\cmidrule(lr){3-4}
\cmidrule(lr){5-6}
& & 4 & 8 & 4 & 8 \\
\hline
\textcolor{gray}{Vanilla} & \textcolor{gray}{-} & \textcolor{gray}{71.8} & \textcolor{gray}{69.4} & \textcolor{gray}{81.6} & \textcolor{gray}{79.0} \\ \hline
\multirow{3}{*}{LoRA}
& 15.1M & 70.6 & 69.0 & 82.1 & 80.9 \\
& 7.6M  & 70.6 & 68.8 & 81.7 & 80.6 \\
& 3.8M  & 70.8 & 69.4 & 81.7 & 80.6 \\ \hline
\multirow{3}{*}{LRR}
& 2.4M & 71.3 & 69.5 & 82.1 & 81.1 \\
& 1.1M & 71.6 & 68.6 & 81.8 & 81.2 \\
& 0.5M & 71.4 & 68.9 & 82.1 & 80.7 \\
\bottomrule
\end{tabular}
\end{table}

Intuitively, requiring $\boldsymbol\Delta_t\rightarrow\boldsymbol0$ in Eq. \ref{eq:method_original_update} can maintain the correctness for an already correct token; thus an ideally trained looped transformer should not have worse results with more loops. However, as illustrated in Fig. \ref{fig:intro_performance_trajectory}(a), the late-stage drift persists at every step. We attribute this drift to an error accumulation feature of the loop block. When the input token is optimal, only when all the looped layers recognize the optimality can the whole block make no interruption, which is hard to achieve since the error is inevitable.

To mitigate this issue, we apply a Latent Residual Rectifier (LRR) after the whole loop block to rectify the error. LRR is a three-layer MLP $\mathcal{R}^\text{LRR}$ applied between two loop iterations in the form of a residual path \cite{he2016deep}, which ensures the training gradient can directly flow through the vanilla path. With LRR, the update can be formulated as:

\begin{equation}
    \widetilde{\boldsymbol\Delta}_t=\boldsymbol\Delta_t+\mathcal R^\text{LRR}(\boldsymbol\Delta_t).
    \label{eq:method_rectified_update}
\end{equation}
\begin{equation}
    \mathbf H_t=
    \mathbf H_{t-1}
    +\boldsymbol\eta_t\odot
    \widetilde{\boldsymbol\Delta}_t.
    \label{eq:method_RDM_LRR_update}
\end{equation}

We compare tuning all the layers using LoRA \cite{hu2022lora} and only rectifying the output hidden states in Table \ref{tab:method_LRR}. The results show that a post-connected reflection outside the loop block brings slightly better results than tuning all the layers, but with much fewer trainable parameters, indicating that all the loop layers should be treated as a whole part.

\begin{algorithm}[t]
\caption{Rectifying Latent Dynamics (RLD)}
\label{alg:rld}
\begin{algorithmic}[1]
\Require Loop block $\mathcal{F}^\text{L-blk.}$,
RDM encoder $\mathcal{E}^\text{RDM}$, RDM decoder $\mathcal{D}^\text{RDM}$, LRR module $ \mathcal{R}^\text{LRR}$, clipping threshold $c$, number of loops $T$,
initial hidden states $\mathbf{H}_0$
\Ensure Final hidden states $\mathbf{H}_T$

\State $\mathbf{S}_0 \gets \mathbf{0}$
\Comment{Initialize depth states}

\For{$t = 1$ to $T$}

    \State $\mathbf{Z}_t\gets\mathcal{F}^{\text{L-blk.}}(\mathbf{H}_{t-1})$
    \Comment{Loop forward}
    
    \State $\boldsymbol\Delta_t\gets\mathbf Z_t-\mathbf H_{t-1}$

    \State $\widetilde{\mathbf{Z}}_t \gets \operatorname{LN}(\mathbf{Z}_t), \quad
    \widetilde{\mathbf{S}}_{t-1} \gets \operatorname{LN}(\mathbf{S}_{t-1})$

    \State $\mathbf{S}_t\gets\mathcal{E}^{\text{RDM}}\left([\widetilde{\mathbf{Z}}_t\Vert \widetilde{\mathbf{S}}_{t-1}]\right) + \mathbf{S}_{t-1}$
    \Comment{Depth update}

    \State $\boldsymbol\eta_t\gets2\sigma\!\left(\mathcal D^\text{RDM}(\mathbf S_t)\right)$

    \State $\widetilde{\boldsymbol\Delta}_t\gets\boldsymbol\Delta_t+\mathcal{R}^\text{LRR}(\boldsymbol\Delta_t)$
    \Comment{LRR forward}

    \For{$i = 1$ to $N$}

        \State $\widehat{\boldsymbol{\Delta}}_{t,i}
        \gets \widetilde{\boldsymbol{\Delta}}_{t,i}
        \cdot \min\left\{
        1,
        \frac{c}{
        \lVert\widetilde{\boldsymbol{\Delta}}_{t,i}
        \rVert_2+\epsilon}
        \right\}$

        \State $\mathbf{H}_{t,i}
        \gets \mathbf{H}_{t-1,i}
        + \eta_{t,i}
        \widehat{\boldsymbol{\Delta}}_{t,i}$
        \Comment{Token update}

    \EndFor

\EndFor

\State \Return $\mathbf{H}_T$
\end{algorithmic}
\end{algorithm}

\subsection{Update Clipping}
\label{sec:method_UC}

\begin{figure}[t]
    \centering
    \includegraphics[width=\linewidth]{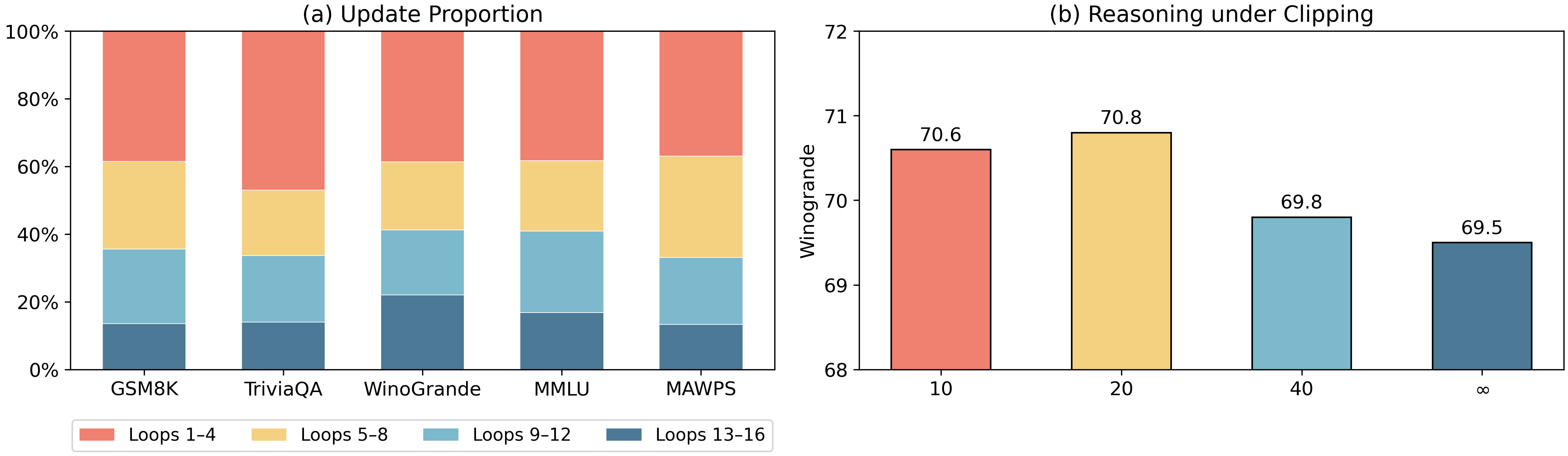}
    \caption{\textbf{(a) Update Proportion:} The L2 distance between adjacent layers in different loop stages. \textbf{(b) Reasoning under Clipping:} Winogrande score for models tuned with update clipping.}
    \label{fig:method_clip}
\end{figure}

We plot the L2 distance of adjacent layers in Fig. \ref{fig:method_clip}(a). In most tasks, the first few loops take up the largest update, while the subsequent steps make less difference. This indicates that the model prefers fast and in-one-go thinking, which does not align with the ideal step-by-step thinking pattern and tends to cause a wrong path \textbf{P1} in Fig. \ref{fig:intro_performance_trajectory} since a few loops are not enough for hard tasks.

Inspired by optimizers adopting gradient clipping \cite{pmlr-v80-shazeer18a} to prevent the model from updating too intensely, we add an update clipping to force the model to think ``more slowly '' and underscore the potential of multi-step reasoning. Let $\widetilde{\boldsymbol\delta}_{t,i} =(\widetilde{\boldsymbol\Delta}_t)_{i,:}^{\mathsf T} \in\mathbb R^d$ denote the rectified residual of token $i$. We clip it along the hidden dimension:

\begin{equation}
    \widehat{\boldsymbol\delta}_{t,i}
    =
    \widetilde{\boldsymbol\delta}_{t,i}
    \min\left\{
        1,\frac{c}{
            \lVert\widetilde{\boldsymbol\delta}_{t,i}\rVert_2
            +\epsilon}
    \right\}.
    \label{eq:method_clipped_update}
\end{equation}

\noindent where $\epsilon$ is a small value to prevent zero-division, and $c>0$ is the update clipping hyperparameter. We choose $c$ for each model by letting around 10\% of tokens be clipped. After clipping, we stack and reassemble $\widehat{\boldsymbol\Delta}_t=[\widehat{\boldsymbol\delta}_{t,1},\ldots,\widehat{\boldsymbol\delta}_{t,N}]^{\mathsf T}$. As shown in Fig. \ref{fig:method_clip}(b), this clipping brings better reasoning ability than following the original update. The complete RLD update with update clipping is

\begin{equation}
    \mathbf H_t = \mathbf H_{t-1}+\boldsymbol\eta_t\odot    \widehat{\boldsymbol\Delta}_t.
    \label{eq:method_final_update}
\end{equation}

\subsection{Training Strategy}
\label{sec:method_train}
\paragraph{Trainable Parameters.} We always keep the model frozen during training, and the only trainable parameters in RLD are $\mathcal{E}^\text{RDM}$, layer normalization scaling parameters, $\mathcal{D}^\text{RDM}$, and $\mathcal{R}^\text{LRR}$. Compared with fine-tuning all the model parameters, this recipe not only reduces the computational burden but also helps the new module focus on token dynamic control.

\paragraph{Training Objective.} The training of RLD adopts a standard Supervised Fine-Tuning (SFT) loss \cite{liu2023visual}. Given the dataset $\mathcal{D}=\{(x_i,y_i)\}_{i=1}^N$ composed of $N$ pairs of prompt $x_i$ and labeled response $y_i$, the loss is formulated as

\begin{equation}
    \mathcal{L}=-\frac{1}{N}\sum_{i=1}^{N}\frac{1}{T_i}\sum_{t=1}^{T_i}\log P_{\theta}\left(y_{i,t} \mid x_i, y_{i,<t}\right),
    \label{eq:train_loss}
\end{equation}

\noindent where $\theta$ denotes the RLD parameters, and $T_i$ denotes number of response tokens. Only the response tokens are supervised.

\paragraph{Module Initialization.} Drawing inspiration from ControlNet \cite{zhang2023adding}, we aim to teach the model to rectify its latent dynamics gradually instead of making a great difference to the sufficiently trained loop block. Therefore, the initialization should keep the model following its original information flow and output logits as much as possible, and also avoid too much zero-initialization that may cause slow convergence \cite{steinwart2019sober}. To this end, we zero-initialize the last layer in $\mathcal{E}^\text{RDM}$ and $\mathcal{R}^\text{LRR}$, along with removing the bias layer in $\mathcal{D}^\text{RDM}$.

\section{Experiment}
\begin{table*}[t]
\centering
\caption{\textbf{Results on LoopVL-1B.} The gray percentages indicate performance relative to the default LoopVL training configuration (H2L3) on each benchmark. ``DS'' denotes the compared method DeepSupervision. The \footnotesize{\textcolor{gray}{downsized number}} denotes the percentage of the default setting.}
\resizebox{\textwidth}{!}{
\begin{tabular}{l|ccccccccc}
\toprule
\textbf{Method}
& \textbf{AI2D} & \textbf{POPE} & \textbf{MME$^\text{P}$} & \textbf{SQA$^\text{IMG}$} & \textbf{InfoVQA} & \textbf{TextVQA} & \textbf{RWQA} & \textbf{VMCB} & \textbf{ChartQA} \\

\rowcolor[HTML]{EFEFEF} \hline
\multicolumn{10}{c}{\textit{Recurrence Depth = H2L3 (Equivalent Depth = 8)}} \\ \hline
LoopVL & \textcolor{gray}{75.3} & \textcolor{gray}{91.7} & \textcolor{gray}{1292} & \textcolor{gray}{73.9} & \textcolor{gray}{60.7} & \textcolor{gray}{53.1} & \textcolor{gray}{70.8} & \textcolor{gray}{70.7} & \textcolor{gray}{76.3} \\
+SFT   & 71.2~\textcolor{gray}{\footnotesize 95\%} & 90.7~\textcolor{gray}{\footnotesize 99\%} & \textbf{1278~\textcolor{gray}{\footnotesize 99\%}} & 53.5~\textcolor{gray}{\footnotesize 72\%} & 55.7~\textcolor{gray}{\footnotesize 92\%} & \textbf{54.0~\textcolor{gray}{\footnotesize 102\%}} & 66.0~\textcolor{gray}{\footnotesize 93\%} & 68.5~\textcolor{gray}{\footnotesize 97\%} & \textbf{74.7~\textcolor{gray}{\footnotesize 98\%}} \\ 
+DS & 67.1~\textcolor{gray}{\footnotesize 89\%} & 89.1~\textcolor{gray}{\footnotesize 97\%} & 1256~\textcolor{gray}{\footnotesize 97\%} & 63.3~\textcolor{gray}{\footnotesize 86\%} & 50.4~\textcolor{gray}{\footnotesize 83\%} & 50.2~\textcolor{gray}{\footnotesize 95\%} & 61.3~\textcolor{gray}{\footnotesize 86\%} & 66.2~\textcolor{gray}{\footnotesize 94\%} & 61.2~\textcolor{gray}{\footnotesize 80\%}\\ \rowcolor[HTML]{DAE3F5}
+RLD   & \textbf{71.4~\textcolor{gray}{\footnotesize 95\%}} & \textbf{91.4~\textcolor{gray}{\footnotesize 100\%}} & 1272~\textcolor{gray}{\footnotesize 98\%} & \textbf{72.2~\textcolor{gray}{\footnotesize 98\%}} & \textbf{59.5~\textcolor{gray}{\footnotesize 98\%}} & 52.0~\textcolor{gray}{\footnotesize 98\%} & \textbf{67.2~\textcolor{gray}{\footnotesize 95\%}} & \textbf{69.7~\textcolor{gray}{\footnotesize 99\%}} & 74.4~\textcolor{gray}{\footnotesize 98\%} \\

\rowcolor[HTML]{EFEFEF} \hline
\multicolumn{10}{c}{\textit{Recurrence Depth = H3L3 (Equivalent Depth = 12)}} \\ \hline
LoopVL & \textcolor{gray}{69.8} & \textcolor{gray}{90.6} & \textcolor{gray}{1162} & \textcolor{gray}{70.3} & \textcolor{gray}{32.0} & \textcolor{gray}{23.3} & \textcolor{gray}{61.4} & \textcolor{gray}{67.8} & \textcolor{gray}{56.8} \\
+SFT   & 71.6~\textcolor{gray}{\footnotesize 95\%} & 89.8~\textcolor{gray}{\footnotesize 98\%} & 1218~\textcolor{gray}{\footnotesize 94\%} & 51.8~\textcolor{gray}{\footnotesize 70\%} & 53.8~\textcolor{gray}{\footnotesize 89\%} & 52.3~\textcolor{gray}{\footnotesize 98\%} & 66.3~\textcolor{gray}{\footnotesize 94\%} & 67.0~\textcolor{gray}{\footnotesize 95\%} & 73.0~\textcolor{gray}{\footnotesize 96\%} \\ 
+DS & 66.6~\textcolor{gray}{\footnotesize 88\%} & 89.4~\textcolor{gray}{\footnotesize 97\%} & 1242~\textcolor{gray}{\footnotesize 96\%} & 56.7~\textcolor{gray}{\footnotesize 77\%} & 48.1~\textcolor{gray}{\footnotesize 79\%} & 47.3~\textcolor{gray}{\footnotesize 89\%} & 60.9~\textcolor{gray}{\footnotesize 86\%} & 65.8~\textcolor{gray}{\footnotesize 93\%} & 61.1~\textcolor{gray}{\footnotesize 80\%} \\ \rowcolor[HTML]{DAE3F5}
+RLD   & \textbf{73.2~\textcolor{gray}{\footnotesize 97\%}} & \textbf{90.2~\textcolor{gray}{\footnotesize 98\%}} & \textbf{1322~\textcolor{gray}{\footnotesize 102\%}} & \textbf{73.4~\textcolor{gray}{\footnotesize 99\%}} & \textbf{58.6~\textcolor{gray}{\footnotesize 97\%}} & \textbf{52.5~\textcolor{gray}{\footnotesize 99\%}} & \textbf{69.3~\textcolor{gray}{\footnotesize 98\%}} & \textbf{69.5~\textcolor{gray}{\footnotesize 98\%}} & \textbf{74.0~\textcolor{gray}{\footnotesize 97\%}} \\

\rowcolor[HTML]{EFEFEF} \hline
\multicolumn{10}{c}{\textit{Recurrence Setting = H4L3 (Equivalent Depth = 16)}} \\ \hline
LoopVL & \textcolor{gray}{6.3}  & \textcolor{gray}{86.6} & \textcolor{gray}{64}   & \textcolor{gray}{11.8} & \textcolor{gray}{7.8}  & \textcolor{gray}{3.0}  & \textcolor{gray}{17.6} & \textcolor{gray}{35.1} & \textcolor{gray}{14.7} \\
+SFT   & 69.9~\textcolor{gray}{\footnotesize 93\%} & 89.0~\textcolor{gray}{\footnotesize 97\%} & 1198~\textcolor{gray}{\footnotesize 93\%} & 48.0~\textcolor{gray}{\footnotesize 65\%} & 48.4~\textcolor{gray}{\footnotesize 80\%} & 45.8~\textcolor{gray}{\footnotesize 86\%} & 64.6~\textcolor{gray}{\footnotesize 91\%} & 65.9~\textcolor{gray}{\footnotesize 93\%} & 65.9~\textcolor{gray}{\footnotesize 86\%} \\ 
+DS & 63.2~\textcolor{gray}{\footnotesize 84\%} & 88.8~\textcolor{gray}{\footnotesize 97\%} & 1204~\textcolor{gray}{\footnotesize 93\%} & 51.6~\textcolor{gray}{\footnotesize 70\%} & 42.9~\textcolor{gray}{\footnotesize 71\%} & 40.7~\textcolor{gray}{\footnotesize 77\%} & 58.8~\textcolor{gray}{\footnotesize 83\%} & 63.9~\textcolor{gray}{\footnotesize 90\%} & 54.6~\textcolor{gray}{\footnotesize 72\%} \\ \rowcolor[HTML]{DAE3F5}
+RLD   & \textbf{72.0~\textcolor{gray}{\footnotesize 96\%}} & \textbf{89.4~\textcolor{gray}{\footnotesize 97\%}} & \textbf{1306~\textcolor{gray}{\footnotesize 101\%}} & \textbf{72.7~\textcolor{gray}{\footnotesize 98\%}} & \textbf{55.7~\textcolor{gray}{\footnotesize 92\%}} & \textbf{49.4~\textcolor{gray}{\footnotesize 93\%}} & \textbf{68.5~\textcolor{gray}{\footnotesize 97\%}} & \textbf{67.2~\textcolor{gray}{\footnotesize 95\%}} & \textbf{72.6~\textcolor{gray}{\footnotesize 95\%}} \\

\bottomrule
\end{tabular}
}
\label{tab:exp_loopvl}
\end{table*}
\subsection{Experiment Settings}
\label{sec:exp_setup}
\paragraph{Models.} We evaluate RLD on both LLMs and VLMs. For LLMs, we adopt the Ouro-1.4B and Ouro-2.6B \cite{zhu2025scaling}, and the early exit gate is disabled during training and evaluation. For VLMs, we choose LoopVL \cite{qian2026loopvl}, which has a standard ViT vision encoder and a looped LLM decoder. The decoder features two loop blocks, H and L, and utilizes a nested looped paradigm. We set two groups of $\mathcal{E}^\text{RDM},\mathcal{D}^\text{RDM},\mathcal{R}^\text{LRR}$ for H and L blocks. Details are in Appendix \ref{sec:supp_model_details}.

\paragraph{Training Data.} We adopt the MetaMathQA-400k dataset \cite{ICLR2024_c400474e} for Ouro-series model training. This dataset contains only math problems and helps compare methods' generalization ability. For LoopVL, we use the DocVQA training set \cite{mathew2021docvqa} and randomly sample 40k text samples from MetaMathQA, forming a multimodal dataset of 80k samples.

\paragraph{Compared Methods.} We compare RLD with standard SFT \cite{wei2022finetuned} and DeepSupervision \cite{chng2026looped} methods. The SFT is implemented by applying LoRA \cite{hu2022lora} and training with the same recipe (depth sampling, training data, optimizer setting, loss function) as RLD. DeepSupervision adds explicit supervision to the logits of each loop iteration and sums them up to formulate the final loss. In our experiments, DeepSupervision shares the same LoRA setting as the SFT group, which is presented in the Appendix \ref{sec:supp_training_details}.

\paragraph{Benchmarks.} We include both text and vision tasks. For text tasks, we evaluate the reasoning (MMLU \cite{hendrycks2021measuring}, GSM8k \cite{cobbe2021training}), conversation understanding (CoQA \cite{reddy-etal-2019-coqa}), and common sense answering (WinoGrande \cite{sakaguchi2021winogrande}, XQuAD \cite{artetxe-etal-2020-cross}). For vision tasks, we choose visual reasoning and understanding (AI2D \cite{10.1007/978-3-319-46493-0_15}, ScienceQA-Image \cite{NEURIPS2022_11332b6b}), OCR and document understanding (InfoVQA \cite{mathew2022infographicvqa}, ChartQA \cite{masry2022chartqa}, TextVQA \cite{singh2019towards}), general and real-world understanding (VMCBench \cite{zhang2025automated}, RealWorldQA \cite{xai2024realworldqa}), and hallucination (POPE \cite{li-etal-2023-evaluating}, MME \cite{NEURIPS2025_d79a27cf}). These benchmarks effectively evaluate methods' comprehensive ability and generalization beyond the math training data. Detailed settings for each benchmark are in Appendix \ref{sec:supp_benchmark_details}.

\paragraph{Implementation Details.} All the experiments are conducted on a machine with 4$\times$ NVIDIA A800-80G GPUs. The update clipping hyperparameter $c$ is set to 20, 24, and 80 for Ouro-1.4B, Ouro-2.6B, and LoopVL, respectively. Following the training recipe from \cite{pmlr-v306-yang26t}, the training depth adopts the same log-normal random loop but with a range of [2, 16]. Evaluations for text and vision tasks are conducted with \texttt{lm-evaluation-harness} \cite{eval-harness} and \texttt{LMMs-Eval} \cite{zhang2024lmmsevalrealitycheckevaluation}. Details on the other training hyperparameters and evaluation settings are presented in Appendix \ref{sec:supp_training_details}.

\subsection{Main Results}
\begin{table*}[htbp]
\centering
\caption{\textbf{Model Performance of Ouro-series.} The vanilla Ouro models are trained with a fixed depth of 4. The number denotes the number of loop iterations during evaluation. The best result of a group is in \textbf{bold}.}
\label{tab:performance_comparison}
\setlength{\tabcolsep}{5pt}
\resizebox{\textwidth}{!}{
\begin{tabular}{l|ccccccccccccccc}
\toprule
\multirow{2}{*}{\textbf{Method}}
& \multicolumn{3}{c}{\textbf{CoQA$^\text{F1}$}}
& \multicolumn{3}{c}{\textbf{WinoGrande$^\text{Acc.}$}}
& \multicolumn{3}{c}{\textbf{MMLU$^\text{Acc.}$}}
& \multicolumn{3}{c}{\textbf{TriviaQA$^\text{Match}$}}
& \multicolumn{3}{c}{\textbf{XQuAD-en$^\text{F1}$}} \\
\cmidrule(lr){2-4}
\cmidrule(lr){5-7}
\cmidrule(lr){8-10}
\cmidrule(lr){11-13}
\cmidrule(lr){14-16}
& \textbf{4} & \textbf{8} & \textbf{16}
& \textbf{4} & \textbf{8} & \textbf{16}
& \textbf{4} & \textbf{8} & \textbf{16}
& \textbf{4} & \textbf{8} & \textbf{16}
& \textbf{4} & \textbf{8} & \textbf{16} \\
\midrule

Ouro-1.4B
& \textcolor{gray}{81.6} & \textcolor{gray}{79.0} & \textcolor{gray}{57.7} & \textcolor{gray}{71.8} & \textcolor{gray}{69.4} & \textcolor{gray}{65.6} & \textcolor{gray}{68.3} & \textcolor{gray}{64.6} & \textcolor{gray}{53.1} & \textcolor{gray}{44.0} & \textcolor{gray}{35.8} & \textcolor{gray}{16.7} & \textcolor{gray}{66.7} & \textcolor{gray}{65.1} & \textcolor{gray}{47.5} \\
~+ SFT
& \textbf{81.8} & 80.8 & 78.0 & 70.1 & 68.6 & 66.6 & \textbf{65.9} & 62.5 & 56.6 & \textbf{44.1} & 38.5 & 30.0 & 67.6 & 64.7 & 54.7 \\
~+ DS
& 81.1 & 80.3 & 77.7 & 68.8 & 67.7 & 66.2 & 62.0 & 62.0 & 57.0 & 42.7 & 37.1 & 29.0 & \textbf{70.7} & 67.4 & 62.3 \\ \rowcolor[HTML]{DAE3F5}
~+ RLD
& 81.3 & \textbf{81.3} & \textbf{81.3} & \textbf{71.0} & \textbf{71.4} & \textbf{72.0} & 65.7 & \textbf{65.9} & \textbf{65.9} & 43.4 & \textbf{43.5} & \textbf{43.4} & 68.8 & \textbf{68.2} & \textbf{68.3} \\

\midrule

Ouro-2.6B
& \textcolor{gray}{83.2} & \textcolor{gray}{83.0} & \textcolor{gray}{60.2} & \textcolor{gray}{76.1} & \textcolor{gray}{75.0} & \textcolor{gray}{72.2} & \textcolor{gray}{73.7} & \textcolor{gray}{70.7} & \textcolor{gray}{65.2} & \textcolor{gray}{53.1} & \textcolor{gray}{48.9} & \textcolor{gray}{35.6} & \textcolor{gray}{66.8} & \textcolor{gray}{65.1} & \textcolor{gray}{42.2} \\
~+ SFT
& 82.4 & 82.1 & 78.4 & 75.5 & 74.5 & 72.4 & 71.2 & 66.8 & 62.2 & 52.6 & 48.7 & 42.3 & 62.0 & 56.9 & 46.3 \\
~+ DS
& 82.6 & 81.8 & 78.6 & \textbf{75.9} & 74.3 & 73.3 & \textbf{71.8} & 68.8 & 64.2 & 51.6 & 47.5 & 40.9 & \textbf{67.3} & 60.8 & 49.7 \\ \rowcolor[HTML]{DAE3F5}
~+ RLD
& \textbf{83.3} & \textbf{83.2} & \textbf{83.2} & 75.3 & \textbf{75.1} & \textbf{75.1} & \textbf{71.8} & \textbf{72.3} & \textbf{72.3} & \textbf{52.7} & \textbf{53.0} & \textbf{53.0} & 67.2 & \textbf{68.4} & \textbf{68.3} \\

\bottomrule
\end{tabular}
}
\end{table*}
\begin{figure*}[t]
    \centering
    \includegraphics[width=\linewidth]{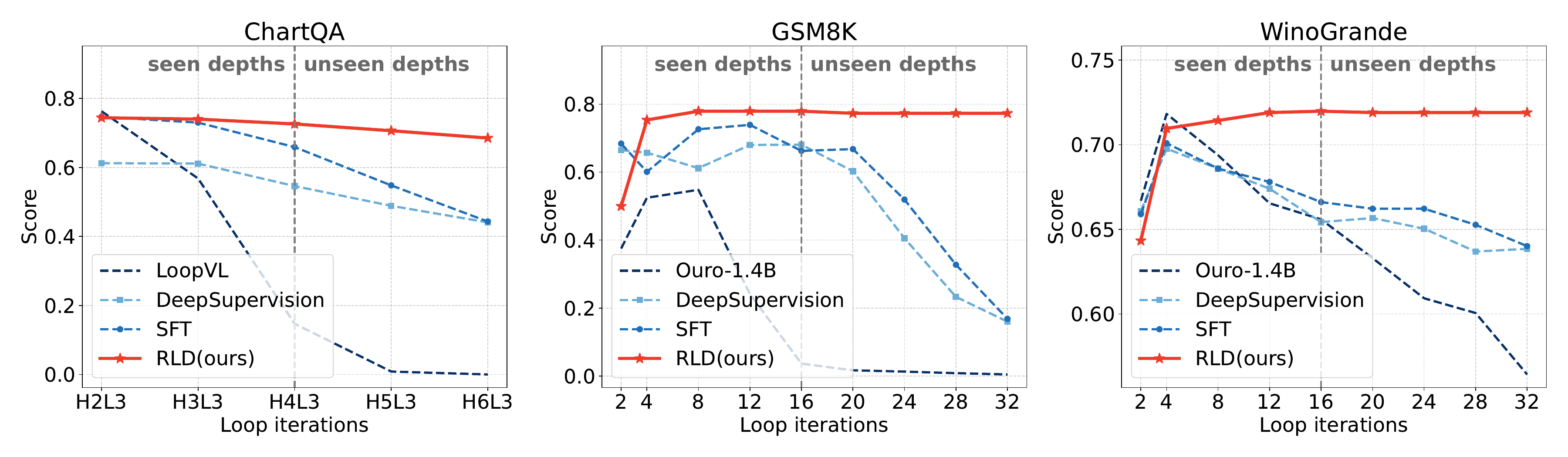}
    \caption{\textbf{Extrapolation Results on LoopVL and Ouro.} The ``seen depths'' and ``unseen depths'' denote the recurrence depths covered and not covered in the training depth sampling distribution.} 
    \label{fig:exp_extrapolation}
\end{figure*}

\paragraph{Evaluations on VLMs.} We first present results on LoopVL in Table \ref{tab:exp_loopvl}. Notably, RLD achieves outstanding stability across almost all tasks and depths and demonstrates superior generalization. On OCR and document understanding tasks, the vanilla model preserves only 19.2\% and 5.7\% of original performance on ChartQA and TextVQA under the H4L3 recurrence setting, while RLD maintains \textbf{95.2\%} and \textbf{93.0\%} for the two tasks with the same setting, surpassing the 86.4\% and 86.3\% results of standard SFT. On MME-Perception, RLD even raises the score from 1272 to 1322 with more loops, unveiling its potential in test-time scaling.

RLD and the compared methods are all trained on multimodal document and math text data. On tasks involving multiple real-world and out-of-domain images, e.g., ScienceQA and RealWorldQA, SFT preserves 65.0\% and 91.1\% of the vanilla model, while RLD maintains \textbf{98.4\%} and \textbf{96.6\%}, respectively. These results support RLD's generalization ability across data and tasks, making it insensitive to training data. We credit this advantage to the fact that RLD focuses on \textit{rectifying} the latent dynamics of tokens, instead of \textit{re-establishing} the model's already-learned representations.

\paragraph{Evaluations on LLMs.} In Table \ref{tab:performance_comparison}, the vanilla model faces a serious performance drop under high loop iterations, as TriviaQA decreases from the highest 44.0 to 16.7. Besides, this issue is more serious on generation-heavy tasks, e.g., GSM8k (Fig. \ref{fig:exp_extrapolation}), which faces a 93.4\% drop from 8 to 16 iterations. SFT and DeepSupervision do \textit{partially mitigate} but do not \textit{fully solve} this issue: SFT achieves 62.7 points higher than the base model at 16, but still faces a 7.7-point drop from its own peak at the loop iteration of 12.

Conversely, RLD achieves outstanding stability across most depth settings, and its advantage is more pronounced under high loop iterations. On the XQuAD task, RLD even brings a 2.1- and 1.6-point boost beyond the best results in the vanilla 1.4B and 2.6B models. As recurrent depth grows from 4 to 16, the performance drop of RLD is consistently $\le$\textbf{0.6} points across all the tasks. When the loop iteration is 8 or 16, RLD outperforms all the compared methods and the base model, demonstrating excellent stability.

\paragraph{Depth Extrapolation.} As shown in Fig. \ref{fig:exp_extrapolation}, RLD successfully extrapolates to unseen depths and achieves 68.5 on ChartQA with H6L3 loops, which is \textbf{54.6\%} higher than SFT. On Ouro, the training data contains only math data, so all the compared methods outperform the vanilla model. However, when the test-time depth exceeds the training depth, both SFT and DeepSupervision face a sharp decrease of $\sim$75\% on GSM8k, while RLD \textbf{barely loses any performance}. On the commonsense reasoning task WinoGrande, compared methods start to degrade after the vanilla model peaks, while RLD grows steadily and stabilizes at the peak value within and beyond the training depth. 

\paragraph{Efficiency Studies.}
\begin{table}[t]
\centering
\caption{\textbf{Training and Inference Efficiency Results.} We report trainable parameter size (Param), wall-clock training time (Time) for training, and Time-To-First-Token (TTFT/ms), Time-Per-Output-Token (TPOT/ms), and max memory (VRAM/GiB) for inference. Test samples are from GSM8k and TextCaps.}
\label{tab:efficiency}
\setlength{\tabcolsep}{3pt}
\renewcommand{\arraystretch}{1.1}
\resizebox{\linewidth}{!}{
\begin{tabular}{l|ccccc}
\toprule
\multirow{2}{*}{\textbf{Method}}
& \multicolumn{2}{c}{\textbf{Training}}
& \multicolumn{3}{c}{\textbf{Inference}} \\
\cmidrule(lr){2-3}
\cmidrule(lr){4-6}
& \textbf{Param$\downarrow$}
& \textbf{Time$\downarrow$}
& \textbf{TTFT$\downarrow$}
& \textbf{TPOT$\downarrow$}
& \textbf{VRAM$\downarrow$} \\
\rowcolor[HTML]{EFEFEF} \hline

\multicolumn{6}{c}{\textit{Ouro-1.4B, Inference Depth = 16}} \\
\hline
SFT
& 15.1M & 11.6h & 297.8\scriptsize{$\pm$0.8} & 302.3\scriptsize{$\pm$0.8} & 3.92 \\
DS
& 15.1M & 12.9h & 298.2\scriptsize{$\pm$0.5} & 301.9\scriptsize{$\pm$0.7} & 3.92 \\
\rowcolor[HTML]{DAE3F5}
RLD
& 4.2M & 8.1h & 309.9\scriptsize{$\pm$2.0} & 306.1\scriptsize{$\pm$1.7} & 3.94 \\

\rowcolor[HTML]{EFEFEF} \hline

\multicolumn{6}{c}{\textit{LoopVL-1B, Inference Depth = H2L3}} \\
\hline
SFT
& 16.5M & 7.3h & 442.0\scriptsize{$\pm$1.7} & 172.6\scriptsize{$\pm$1.1} & 6.41 \\
DS
& 16.5M & 7.9h & 442.3\scriptsize{$\pm$0.9} & 172.9\scriptsize{$\pm$1.2} & 6.41 \\
\rowcolor[HTML]{DAE3F5}
RLD
& 4.9M & 4.0h & 454.8\scriptsize{$\pm$1.1} & 180.8\scriptsize{$\pm$0.8} & 6.43 \\

\bottomrule
\end{tabular}
}
\end{table}

As shown in Table \ref{tab:efficiency}, RLD effectively reduces parameter size and time consumption during training, and introduces only little inference overhead. On the LoopVL model, RLD reduces \textbf{45.2\%} wall-clock time with only \textbf{29.7\%} trainable parameters compared with standard SFT, while maintaining the additional prefill latency lower than 3\%. On Ouro, RLD training consumes less than \textbf{62.8\%} of DeepSupervision, and the inference overhead are 3.9\% and 1.4\% for prefill and decode, respectively. For both models, the extra memory is 0.02 GiB, which is trivial during practical deployment. These results demonstrate the high training efficiency and low inference overhead of RLD, making its implementation ideal and lightweight.

\subsection{Ablation Studies}
\begin{table}[t]
\centering
\caption{\textbf{Ablation Studies on RDM and LRR.} The first group are results of GSM8k on Ouro-1.4B, and the second group shows results on ChartQA on LoopVL. The Update Clipping is disabled.}
\label{tab:ablation}
\resizebox{\linewidth}{!}{
\begin{tabular}{cc|ccc|cc}
\toprule
\multicolumn{2}{c|}{\textbf{Method}}
& \multicolumn{3}{c|}{\textbf{Seen Depths}}
& \multicolumn{2}{c}{\textbf{Unseen Depths}} \\
\cmidrule(lr){1-2}
\cmidrule(lr){3-5}
\cmidrule(lr){6-7}
\textbf{RDM}
& \textbf{LRR}
& \textbf{4}
& \textbf{8}
& \textbf{16}
& \textbf{24}
& \textbf{32} \\
\midrule
\ding{51} & & 45.6 & 44.4 & 42.8 & 40.9 & 40.4 \\
& \ding{51} & 50.7 & 54.1 & 5.2 & 1.1 & 0.5 \\ 
\rowcolor[HTML]{DAE3F5}
\ding{51} & \ding{51} & 72.1 & 72.3 & 72.2 & 73.0 & 73.0 \\

\midrule

& & \textbf{H2L3} & \textbf{H3L3} & \textbf{H4L3} & \textbf{H5L3} & \textbf{H6L3} \\ \midrule
\ding{51} & & 72.4 & 72.2 & 70.1 & 68.1 & 65.6 \\
& \ding{51} & 73.5 & 73.0 & 66.7 & 52.8 & 38.4 \\ \rowcolor[HTML]{DAE3F5}
\ding{51} & \ding{51} & 74.2 & 74.3 & 73.2 & 71.2 & 68.2 \\

\bottomrule
\end{tabular}
}
\end{table}

\begin{figure}[t]
    \centering
    \includegraphics[width=\linewidth]{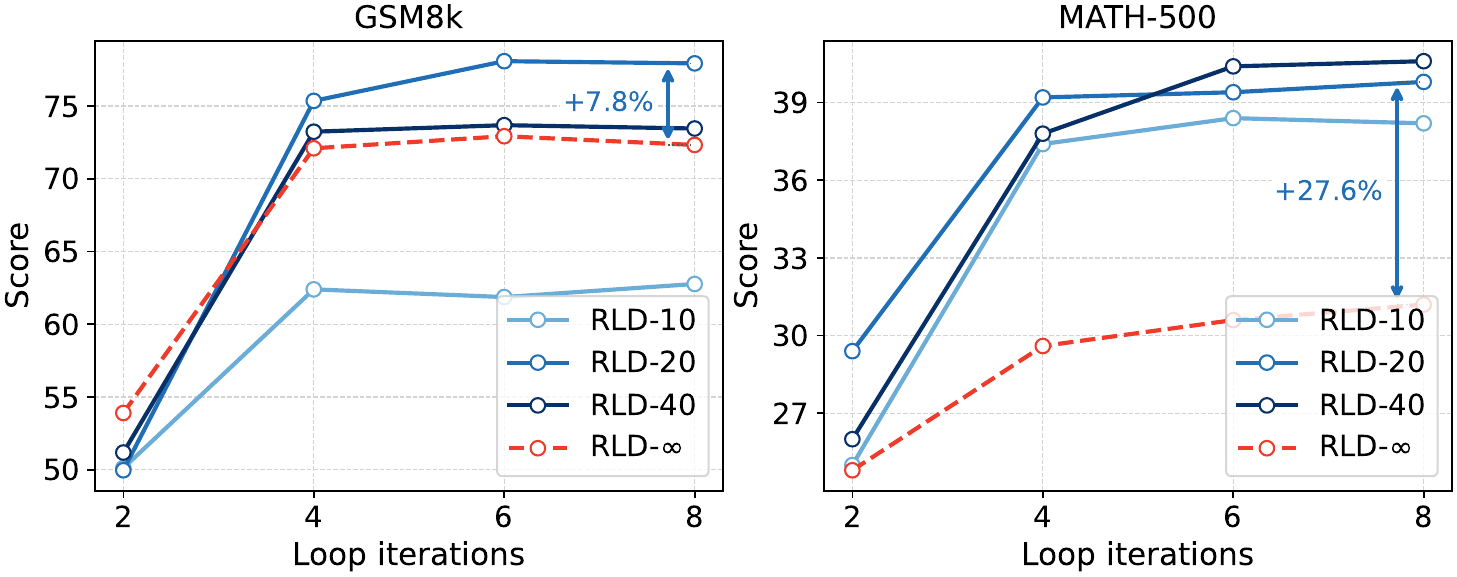}
    \caption{\textbf{Ablation Study on Update Clipping.} We retrain RLD with different update clipping parameter $c$. Smaller $c$ has more clipping, and ``RLD-$\infty$'' equals to fully removing Update Clipping}
    \label{fig:exp_ablation_c}
\end{figure}

\paragraph{Ablation on RDM and LRR.}

Results on ablating RDM and LRR are presented in Table \ref{tab:ablation}. The trend on Ouro and LoopVL is similar: RDM alone provides stable performance across depths but yields lower accuracy; LRR achieves higher scores at shallow depths but degrades as depth increases. We hypothesize that this is attributed to RDM changing the magnitude for stability; however, the hidden space of looped transformers is nonlinear, and the direction is coupled with magnitude and should be adjusted at the same time, which is exactly the role of LRR. Combining RDM and LRR therefore achieves a better balance between performance and stability than either component alone.

\paragraph{Ablation on Update Clipping.}
As illustrated in Fig. \ref{fig:exp_ablation_c}, we plot the model's math reasoning performance with different update clipping parameter $c$. We find that restricting the update magnitude brings significant benefit when the reasoning depth is high, e.g., a 27.6\% relative boost on MATH-500. These results highlight that Update Clipping effectively underscores the reasoning potential of looped transformers, and supports our manual ``slow down'' operation.

\subsection{Discussions}
\paragraph{Depth Settings in LoopVL.}
\begin{table}[t]
\centering
\caption{\textbf{RLD Results on LoopVL under More Depths.} The number below each setting denotes the equivalent depth. The \textcolor{gray}{gray} number denotes the percentage of the optimal results.}
\label{tab:exp_loopvl_more_depths}
\setlength{\tabcolsep}{2pt}
\renewcommand{\arraystretch}{1.05}
\resizebox{\linewidth}{!}{
\begin{tabular}{c|ccccc}
\toprule
\multirow{2}{*}{\textbf{Task}}
& \textbf{H3L2}
& \textbf{H4L2}
& \textbf{H3L3}
& \textbf{H3L4}
& \textbf{H4L4} \\
\cmidrule(lr){2-6}
& \textbf{9}
& \textbf{12}
& \textbf{12}
& \textbf{15}
& \textbf{20} \\
\midrule

AI2D
& 71.5~\textcolor{gray}{\footnotesize 98\%} & 72.2~\textcolor{gray}{\footnotesize 99\%} & 73.2~\textcolor{gray}{\footnotesize 100\%} & 72.4~\textcolor{gray}{\footnotesize 99\%} & 71.4~\textcolor{gray}{\footnotesize 98\%} \\

ChartQA
& 73.6~\textcolor{gray}{\footnotesize 99\%} & 72.8~\textcolor{gray}{\footnotesize 98\%} & 74.0~\textcolor{gray}{\footnotesize 100\%} & 74.1~\textcolor{gray}{\footnotesize 100\%} & 72.0~\textcolor{gray}{\footnotesize 97\%} \\

MME
& 1299~\textcolor{gray}{\footnotesize 98\%} & 1283~\textcolor{gray}{\footnotesize 97\%} & 1322~\textcolor{gray}{\footnotesize 100\%} & 1308~\textcolor{gray}{\footnotesize 99\%} & 1286~\textcolor{gray}{\footnotesize 97\%} \\

POPE
& 89.8~\textcolor{gray}{\footnotesize 100\%} & 89.8~\textcolor{gray}{\footnotesize 100\%} & 90.2~\textcolor{gray}{\footnotesize 100\%} & 89.6~\textcolor{gray}{\footnotesize 99\%} & 89.0~\textcolor{gray}{\footnotesize 99\%} \\

\bottomrule
\end{tabular}
}
\end{table}

Apart from fixing the \texttt{L} loop to 3 in Table \ref{tab:exp_loopvl}, we further study adjusting the recurrent depth by changing L and present the results of RLD in Table \ref{tab:exp_loopvl_more_depths}. The optimal depth setting still prefers L to be 3, since the base model is sufficiently trained under this setting. However, it is worth noting that the performance difference between settings is consistently lower than 3\% when the depth grows from 9 to 20. These results show RLD's robustness for models with a complex or hierarchical architecture.

\paragraph{Attention Visualization.}
\begin{figure}[t]
    \centering
    \includegraphics[width=0.95\linewidth]{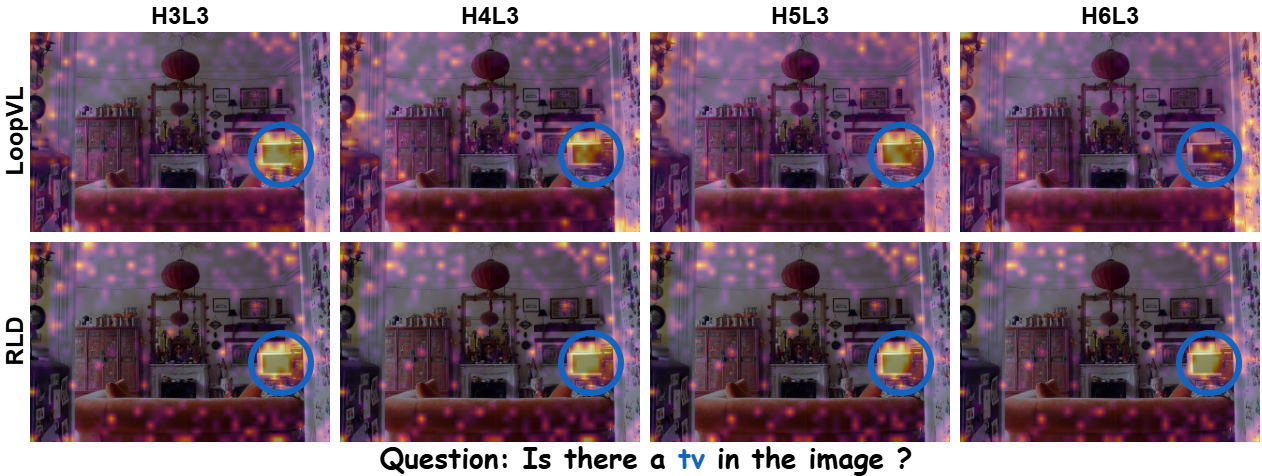}
    \caption{\textbf{Attention Visualization on LoopVL and RLD.} We plot the average attention map of the \texttt{H} block in LoopVL. The \textcolor[HTML]{1464BF}{blue} circle denotes the key region corresponding to the prompt.}
    \label{fig:attn_vis}
\end{figure}

As illustrated in Fig. \ref{fig:attn_vis}, we plot the attention distribution of LoopVL under different loop settings. As the loop iteration grows, the distribution of the vanilla model disperses and gradually defocuses from the key region of the question, i.e., the television in this example. In contrast, the attention of RLD consistently concentrates on the right place, even when the loop iteration reaches H6L3. More examples are provided in Appendix \ref{sec:supp_attention_map}.

\begin{figure}[t]
    \centering
    \includegraphics[width=\linewidth]{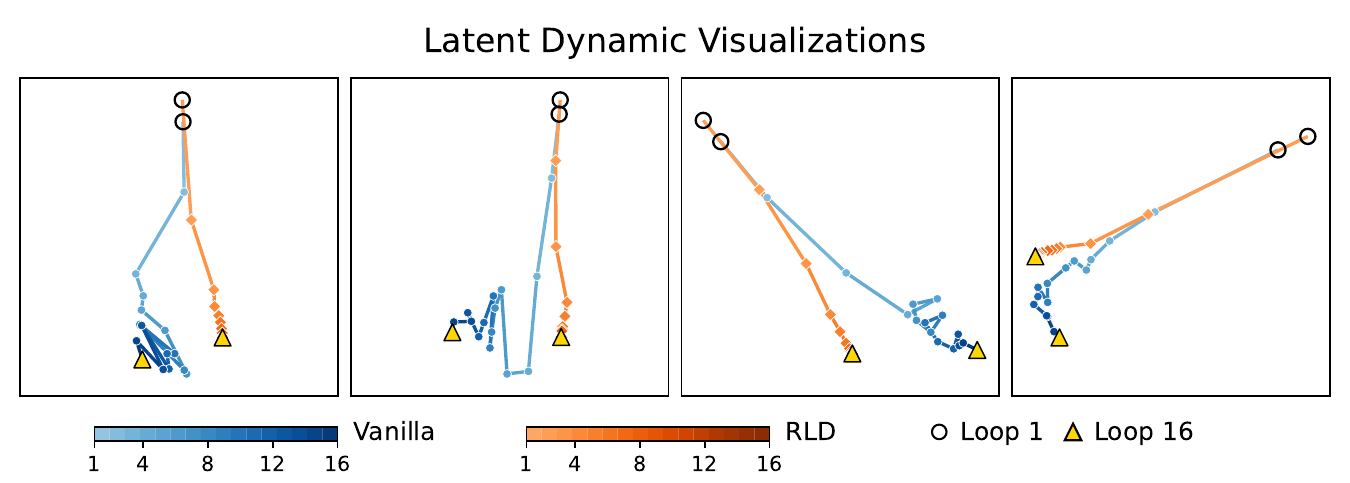}
    \caption{\textbf{Token-Level Visualizations on Vanilla and Rectified Latent Dynamics.} The loop iteration for both models is set to 16. More examples are provided in Appendix \ref{sec:supp_vis_latent_dynamics}.}
    \label{fig:exp_rectified_traces}
\end{figure}

\paragraph{Latent Dynamic Visualizations.}
As shown in Fig. \ref{fig:exp_rectified_traces}, most tokens move randomly and disorderedly in the late stage in the vanilla model; this drift consistently pushes the token away and makes it unable to stabilize in a correct region, causing the \textbf{P2} in Fig. \ref{fig:intro_performance_trajectory}. Conversely, the trace of RLD exhibits a smoother and more consistent trend, demonstrating much less late-stage drift. This qualitative visualization supports the effectiveness of RLD, which makes the latent dynamics of looped transformers more controllable.

\paragraph{Update Magnitude Visualizations.}

\begin{figure}[t]
    \centering
    \includegraphics[width=0.8\linewidth]{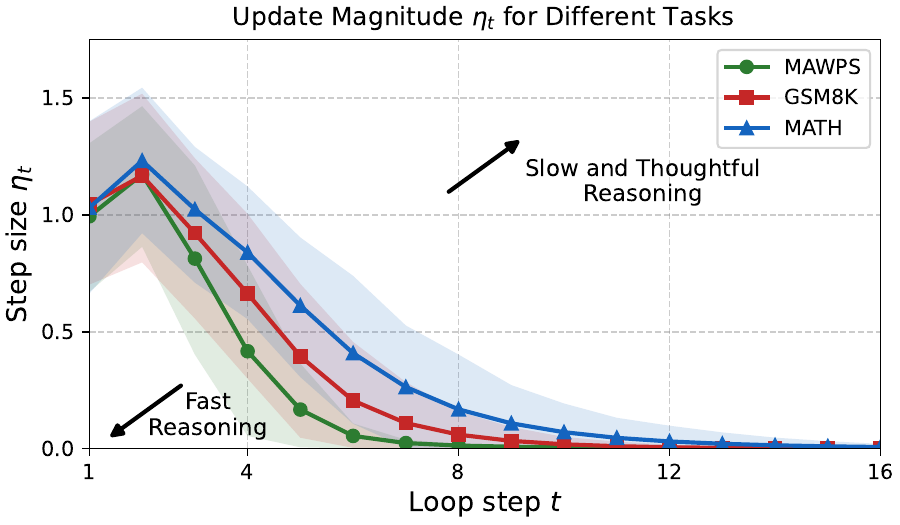}
    \caption{\textbf{Visualization of Update Magnitude with Loop Iterations.} The shaded region represents the envelope containing the 90\% trajectories closest to the mean curve.}
    \label{fig:exp_eta_vis}
\end{figure}

The RDM module controls how much of the thinking should be accepted at each step. We choose three datasets that represent simple algebra, primary school math, and competition level, respectively. As shown in Fig. \ref{fig:exp_eta_vis}, RLD tends to assign higher weights to harder problems to enable deep, slow thinking, while assigning smaller weights to simpler questions to achieve fast convergence. Besides, as discussed in Appendix \ref{sec:supp_adaptive_update_scheduling}, \textit{the key advantage of RDM is not simply stopping the late-stage update, but deciding whether and when to stop dynamically and adaptively to fully release the model capability.}

\section{Conclusion}

In this paper, we analyse the latent dynamics inside looped transformers and find that the preferred fast thinking and late-stage drift cause the overthinking problem. To stabilize the model performance under high loop iterations, we propose a Recurrent Depth Modulator to inject task-relevant explicit depth signal and control the update magnitude, a Latent Residual Rectifier to adjust the evolution direction, and an Update Clipping policy to force the model to slow down thinking and underscore its reasoning potential. Combining these three modules, our method effectively stabilizes the model under high loop iterations and generalizes to unseen tasks and depths. Beyond the performance gain, RLD keeps the base model frozen and is fast to train, making it a promising way to improve looped transformers.

\section*{LLM Usage Statement}
OpenAI Codex is only used to assist with the implementation of research code. The manuscript text and figures are prepared by the authors without generative AI assistance. The authors have reviewed the code and take full responsibility for the implementation, experiments, and manuscript.

{
    \small
    \bibliographystyle{ieeenat_fullname}
    \bibliography{main}
}

\clearpage
\setcounter{page}{1}
\maketitlesupplementary

\section{Adaptive Update Scheduling}
\label{sec:supp_adaptive_update_scheduling}
\subsection{Ablation on RDM Paths}
\begin{table}[h]
\centering
\caption{\textbf{Ablation Studies on RDM Paths.} All the ablations are re-trained with the same setting as the main experiment.}
\label{tab:supp_ablation}
\resizebox{\linewidth}{!}{
\begin{tabular}{cc|ccc|cc}
\toprule
\multicolumn{2}{c|}{\textbf{State Access}}
& \multicolumn{3}{c|}{\textbf{Seen Depths}}
& \multicolumn{2}{c}{\textbf{Unseen Depths}} \\
\cmidrule(lr){1-2}
\cmidrule(lr){3-5}
\cmidrule(lr){6-7}
\textbf{Token}
& \textbf{Depth}
& \textbf{4}
& \textbf{8}
& \textbf{16}
& \textbf{24}
& \textbf{32} \\

\rowcolor[HTML]{EFEFEF} \hline
\multicolumn{7}{c}{\textit{MMLU}} \\
\hline
\ding{51} & & 65.1 & 65.2 & 64.4 & 63.8 & 62.7 \\ 
& \ding{51} & 65.5 & 65.6 & 65.6 & 65.5 & 65.5 \\
\rowcolor[HTML]{DAE3F5}
\ding{51} & \ding{51} & 65.7 & 65.9 & 65.9 & 66.1 & 66.1 \\ 

\bottomrule
\end{tabular}
}
\end{table}

In our full implementation, the RDM module controls the token update based on both depth and token states. The depth state injects an iteration-loop perception, and the token state provides task-aware difficulty information. We quantitatively evaluate the function of the two paths in Table \ref{tab:supp_ablation} by re-training the model with RDM access to one single path. The results show that the depth state mainly accounts for stability, as the MMLU score from 4 to 16 changes by only 0.1. However, relying solely on the depth state creates a static decay policy that stops model updates in the late stage and cannot adapt to token difficulty. When combining both depth and token states, the model achieves higher accuracy results and demonstrates relatively steady performance. These results highlight RDM's advantage over static update stopping methods, formulating a task-adaptive update scheduler.

\subsection{Task-Adaptive Stabilization across Depths}
\begin{figure}[h]
    \centering
    \includegraphics[width=\linewidth]{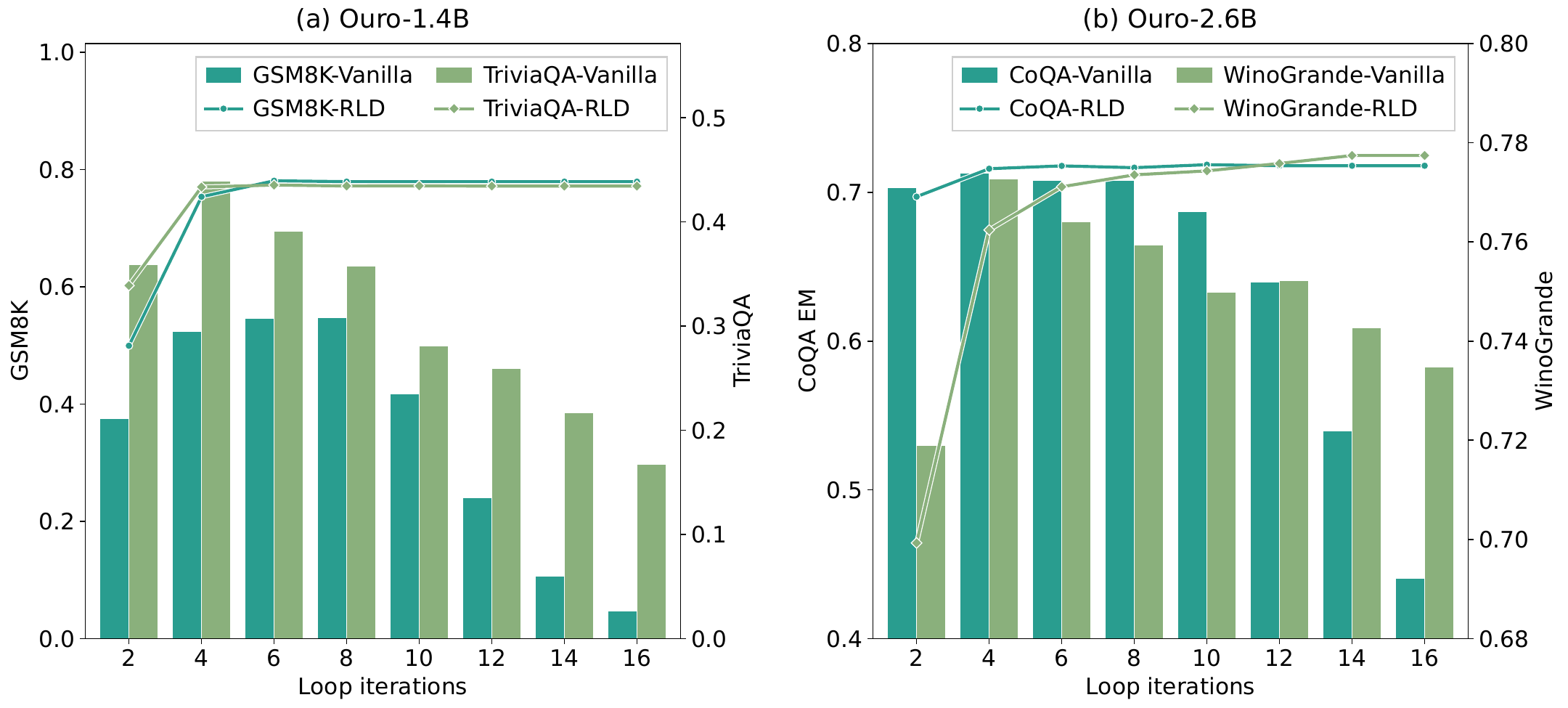}
    \caption{\textbf{Model Performance with Depth Sweep.} RLD effectively maintains the accuracy at the depth where the vanilla model is optimal for different task types.}
    \label{fig:supp_depth_sweep}
\end{figure}
To investigate whether RLD achieves depth robustness through task-adaptive regulation rather than uniformly suppressing late-stage updates, we compare the performance of RLD and the vanilla model across multiple recurrent iterations in Fig. \ref{fig:supp_depth_sweep}. The vanilla model exhibits distinct performance peaks across tasks: GSM8K continues to benefit from loops until approximately 8 loops, whereas CoQA reaches its peak at around 4 loops. RLD preserves these task-dependent improvement patterns while mitigating the subsequent performance degradation, stabilizing near 8 loops on GSM8K and 4 loops on CoQA, respectively. This task-dependent onset of stabilization suggests that \textbf{RLD does not enforce a universal effective computation depth, but instead adapts the progression of latent updates to the computational demands of different tasks.} Such behavior is consistent with the state-conditioned, token-wise update regulation of RDM, allowing useful intermediate refinements to be retained while limiting detrimental updates at later stages.

\section{Experiment Details}
\subsection{Model Details}
\label{sec:supp_model_details}

\begin{figure}[t]
    \centering
    \includegraphics[width=\linewidth]{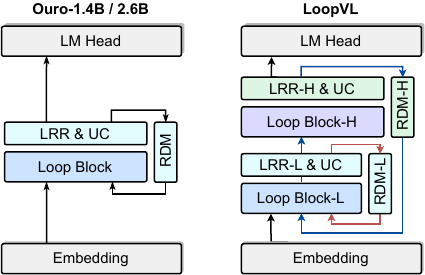}
    \caption{\textbf{RLD Implementation on Ouro and LoopVL.} Ouro features only one set of loop blocks, while LoopVL features one ``L'' set and ``H'' set. For LoopVL, we set two sets of RLD modules for ``L'' and ``H'', respectively.}
    \label{fig:supp_model_implementation}
\end{figure}

We illustrate the RLD implementation for Ouro and LoopVL in Fig. \ref{fig:supp_model_implementation}. The Ouro model adopts a straightforward and simple architecture: all the transformer layers inside the loop block are stacked sequentially, and the number of calls for each layer is equal. We set only one group of $\mathcal{E}^\text{RDM}, \mathcal{D}^\text{RDM}, \mathcal{R}^\text{LRR}$ for Ouro-1.4B and Ouro-2.6B. 

LoopVL is a LLaVA-style \cite{liu2023visual} VLM with a standard ViT vision encoder and a looped LLM decoder. The LLM decoder adopts a nested loop paradigm: there are two sets of loop blocks in LoopVL, i.e., H and L. The L blocks are looped inside H loops, for instance, an H2L3 setting calls the blocks by \texttt{LLLHLLLH}, forming an equivalent recurrent depth of 8. In our implementation, we set two groups of RDM and LRR parameters for the two blocks, and the two blocks share a same clipping threshold $c$.

\subsection{Benchmark Details}
\label{sec:supp_benchmark_details}

Our evaluations are conducted with the \texttt{lm-evaluation-harness v0.4.13} \cite{eval-harness} and \texttt{LMMs-Eval v0.7.4} \cite{zhang2024lmmsevalrealitycheckevaluation}. We have presented the task key names in Table \ref{tab:supp_evaluation_settings}. All multimodal benchmarks are evaluated in a zero-shot setting. For text-only benchmarks, we adopt five-shot evaluation for WinoGrande and TriviaQA, while CoQA, MMLU, and XQuAD are evaluated in a zero-shot setting. For GSM8K, we employ zero-shot chain-of-thought prompting of ``Let's think step by step." We follow the benchmark-specific evaluation protocols and metrics implemented in the respective evaluation harnesses, ensuring consistent evaluation settings across all methods.


\begin{table}[t]
\centering
\caption{\textbf{Training Hyperparameters for Two Models.} Paired values in the Ouro column correspond to the 1.4B and 2.6B variants.}
\label{tab:supp_lora_sft_hyperparameters}
\small
\setlength{\tabcolsep}{6pt}
\begin{tabular}{lcc}
\toprule
\textbf{Hyperparameter} & \textbf{Ouro-1.4B / 2.6B} & \textbf{LoopVL} \\
\rowcolor[HTML]{EFEFEF} \hline
\multicolumn{3}{c}{\textit{LoRA Settings}} \\
\hline
Rank $r$ & \multicolumn{2}{c}{16} \\
Scaling parameter $\alpha$ & \multicolumn{2}{c}{32} \\
Scaling factor $\alpha/r$ & \multicolumn{2}{c}{2} \\
Dropout & 0.05 & 0 \\
Trainable parameters & 15.1M / 30.2M & 16.5M \\
\rowcolor[HTML]{EFEFEF} \hline
\multicolumn{3}{c}{\textit{Training Settings}} \\
\hline
Optimizer & \multicolumn{2}{c}{AdamW} \\
Learning rate & \multicolumn{2}{c}{$1\times10^{-4}$} \\
Adam $(\beta_1,\beta_2)$ & \multicolumn{2}{c}{$(0.9,0.999)$} \\
Adam $\epsilon$ & \multicolumn{2}{c}{$1\times10^{-8}$} \\
Weight decay & \multicolumn{2}{c}{0} \\
Learning-rate schedule & \multicolumn{2}{c}{Cosine decay} \\
Warmup ratio & \multicolumn{2}{c}{0.03} \\
Max gradient norm & \multicolumn{2}{c}{1.0} \\
Epochs & \multicolumn{2}{c}{1} \\
Global batch size & \multicolumn{2}{c}{32} \\
\rowcolor[HTML]{EFEFEF} \hline
\multicolumn{3}{c}{\textit{Recurrence Settings}} \\
\hline
Depth sampling & \multicolumn{2}{c}{Log normal} \\
Sampling mean $\mu$ & 1.7 & 2.0 \\
Sampling std $\sigma$ & \multicolumn{2}{c}{0.4} \\
Depth state dimension $d_s$ & \multicolumn{2}{c}{256} \\
\rowcolor[HTML]{EFEFEF} \hline
\multicolumn{3}{c}{\textit{Data Settings}} \\
\hline
Text data & 400k MetaMath & 40k MetaMath \\
Multimodal data & - & 40k DocVQA \\
\bottomrule
\end{tabular}

\end{table}
\begin{table*}[t]
\centering
\caption{\textbf{Evaluation Settings for Different Benchmarks.} 
Evaluation harnesses, task keys, numbers of few-shot examples, 
and evaluation metrics used in our experiments.}
\label{tab:supp_evaluation_settings}
\begin{tabular}{l|lccc}
\toprule
\textbf{Task}
& \textbf{Evaluation Harness}
& \textbf{Key}
& \textbf{Shots}
& \textbf{Metric} \\
\midrule

AI2D       & LMMs-Eval & ai2d          & 0 & exact\_match \\
POPE       & LMMs-Eval & pope          & 0 & pope\_f1\_score \\
MME        & LMMs-Eval & mme           & 0 & mme\_perception\_score \\
SQA        & LMMs-Eval & scienceqa\_img & 0 & exact\_match \\
InfoVQA    & LMMs-Eval & infovqa\_val   & 0 & anls \\
TextVQA    & LMMs-Eval & textvqa\_val   & 0 & exact\_match \\
RWQA       & LMMs-Eval & realworldqa   & 0 & exact\_match \\
VMCB       & LMMs-Eval & vmcbench      & 0 & average \\
ChartQA    & LMMs-Eval & chartqa       & 0 & relaxed\_overall \\

\midrule

CoQA       & LM Evaluation Harness & coqa               & 0 & f1 \\
WinoGrande & LM Evaluation Harness & winogrande         & 5 & acc \\
MMLU       & LM Evaluation Harness & mmlu               & 0 & acc \\
TriviaQA   & LM Evaluation Harness & triviaqa           & 5 & exact\_match \\
XQuAD      & LM Evaluation Harness & xquad\_en          & 0 & f1 \\
GSM8k      & LM Evaluation Harness & gsm8k\_cot\_zeroshot & 0 & exact\_match (flexible-extract) \\
MATH-500   & LM Evaluation Harness & minerva\_math500 & 4 & math\_verify \\

\bottomrule
\end{tabular}
\end{table*}

\subsection{Training Details}
\label{sec:supp_training_details}

We have presented the training hyperparameters and LoRA settings in Table \ref{tab:supp_lora_sft_hyperparameters}. In our implementations, all three methods, including RLD, share the same training, recurrence, and data settings. We set the maximum learning rate to $1\times10^{-4}$, and train with a global batch size of 32 with 1 epoch for all three evaluated models. Compared with Ouro, the depth sampling mean $\mu$ is slightly larger since the vanilla LoopVL model is trained with an equivalent depth of 8, while the vanilla Ouro is trained with a depth of 4.

\section{Extended Experiments}
\subsection{Comparison with STARS}
\begin{table*}[htbp]
\centering
\caption{\textbf{Comparison with STARS on Ouro-1.4B.}
We reproduce STARS using the same training dataset as RLD and the LoRA configuration in Table \ref{tab:supp_lora_sft_hyperparameters}, while following the original STARS paper for the
learning rate.}
\label{tab:supp_stars}
\setlength{\tabcolsep}{5pt}
\resizebox{\textwidth}{!}{
\begin{tabular}{l|ccccccccccccccc}
\toprule
\multirow{2}{*}{\textbf{Method}}
& \multicolumn{3}{c}{\textbf{CoQA$^\text{F1}$}}
& \multicolumn{3}{c}{\textbf{WinoGrande$^\text{Acc.}$}}
& \multicolumn{3}{c}{\textbf{MMLU$^\text{Acc.}$}}
& \multicolumn{3}{c}{\textbf{GSM8k$^\text{EM.}$}}
& \multicolumn{3}{c}{\textbf{XQuAD-en$^\text{F1}$}} \\
\cmidrule(lr){2-4}
\cmidrule(lr){5-7}
\cmidrule(lr){8-10}
\cmidrule(lr){11-13}
\cmidrule(lr){14-16}
& \textbf{4} & \textbf{8} & \textbf{16}
& \textbf{4} & \textbf{8} & \textbf{16}
& \textbf{4} & \textbf{8} & \textbf{16}
& \textbf{4} & \textbf{8} & \textbf{16}
& \textbf{4} & \textbf{8} & \textbf{16} \\
\midrule

Ouro-1.4B
& \textcolor{gray}{81.6} & \textcolor{gray}{79.0} & \textcolor{gray}{57.7} & \textcolor{gray}{71.8} & \textcolor{gray}{69.4} & \textcolor{gray}{65.6} & \textcolor{gray}{68.3} & \textcolor{gray}{64.6} & \textcolor{gray}{53.1} & \textcolor{gray}{52.5} & \textcolor{gray}{54.8} & \textcolor{gray}{3.6} & \textcolor{gray}{66.7} & \textcolor{gray}{65.1} & \textcolor{gray}{47.5} \\
~+ STARS & 79.7 & 76.0 & 73.0 & 64.3 & 61.9 & 62.4 & 48.3 & 39.0 & 37.4 & 72.9 & 71.0 & 69.7 & 66.1 & 59.6 & 46.6 \\ \rowcolor[HTML]{DAE3F5}
~+ RLD
& \textbf{81.3} & \textbf{81.3} & \textbf{81.3} & \textbf{71.0} & \textbf{71.4} & \textbf{72.0} & \textbf{65.7} & \textbf{65.9} & \textbf{65.9} & \textbf{75.4} & \textbf{77.9} & \textbf{77.9} & \textbf{68.8} & \textbf{68.2} & \textbf{68.3} \\

\bottomrule
\end{tabular}
}
\end{table*}
We further compare RLD with STARS \cite{pmlr-v306-yang26t}, a closely related approach that stabilizes recurrent dynamics through Jacobian spectral radius regularization. As official fine-tuned checkpoints are unavailable and the released implementation is incomplete, we reproduce STARS on Ouro-1.4B ourselves. STARS requires \textbf{23.5 hours} of training, approximately 2.9$\times$ that of RLD. As shown in Table \ref{tab:supp_stars}, RLD consistently outperforms STARS across all five benchmarks and three evaluated recurrence depths. Although STARS effectively mitigates performance degradation on GSM8K, it exhibits substantial performance drops on several non-math and mixed-domain benchmarks, especially MMLU and WinoGrande. These results demonstrate that RLD achieves a more favorable balance between recurrence-depth stability and cross-task generalization while requiring substantially less training time.

\subsection{Benchmark Results}
\begin{table*}[t]
\centering
\caption{\textbf{MMLU Results by Subject Group.} Zero-shot accuracy (\%) on the four MMLU subject groups. }
\label{tab:supp_mmlu_groups}

\begin{tabular}{l|cccccccccccc}
\toprule
\multirow{2}{*}{\textbf{Method}}
 & \multicolumn{3}{c}{\textbf{Humanities}} & \multicolumn{3}{c}{\textbf{STEM}} & \multicolumn{3}{c}{\textbf{Social Sciences}} & \multicolumn{3}{c}{\textbf{Other}} \\
\cmidrule(lr){2-4}\cmidrule(lr){5-7}\cmidrule(lr){8-10}\cmidrule(lr){11-13}
 & \textbf{4} & \textbf{8} & \textbf{16} & \textbf{4} & \textbf{8} & \textbf{16} & \textbf{4} & \textbf{8} & \textbf{16} & \textbf{4} & \textbf{8} & \textbf{16} \\
\midrule
Ouro-1.4B & \textcolor{gray}{63.8} & \textcolor{gray}{61.0} & \textcolor{gray}{49.1} & \textcolor{gray}{66.8} & \textcolor{gray}{61.2} & \textcolor{gray}{46.1} & \textcolor{gray}{75.3} & \textcolor{gray}{71.8} & \textcolor{gray}{61.6} & \textcolor{gray}{69.7} & \textcolor{gray}{66.3} & \textcolor{gray}{58.1} \\
~+ SFT & \textbf{62.0} & 59.4 & 53.0 & 61.9 & 57.0 & 52.0 & \textbf{74.1} & 70.9 & 64.3 & \textbf{67.5} & 64.2 & 59.1 \\
~+ DS & 58.0 & 58.0 & 52.7 & 57.8 & 57.8 & 52.5 & 70.1 & 70.1 & 66.0 & 64.3 & 64.3 & 59.0 \\
\rowcolor[HTML]{DAE3F5}
~+ RLD & 61.3 & \textbf{61.6} & \textbf{61.6} & \textbf{64.7} & \textbf{65.0} & \textbf{65.0} & 71.9 & \textbf{72.2} & \textbf{72.2} & 67.1 & \textbf{67.1} & \textbf{67.1} \\
\midrule
Ouro-2.6B & \textcolor{gray}{69.2} & \textcolor{gray}{66.5} & \textcolor{gray}{61.2} & \textcolor{gray}{72.6} & \textcolor{gray}{69.0} & \textcolor{gray}{62.8} & \textcolor{gray}{80.1} & \textcolor{gray}{78.0} & \textcolor{gray}{71.2} & \textcolor{gray}{75.3} & \textcolor{gray}{71.8} & \textcolor{gray}{67.6} \\
~+ SFT & \textbf{67.5} & 64.3 & 59.1 & 65.9 & 58.7 & 52.3 & 79.3 & 75.7 & 72.8 & 73.9 & 70.0 & 66.4 \\
~+ DS & 67.3 & 64.4 & 60.2 & 68.7 & 63.5 & 55.2 & \textbf{79.8} & 77.5 & 74.7 & \textbf{74.0} & 72.0 & 69.1 \\
\rowcolor[HTML]{DAE3F5}
~+ RLD & 67.0 & \textbf{66.8} & \textbf{66.7} & \textbf{70.8} & \textbf{71.5} & \textbf{71.6} & 78.5 & \textbf{79.6} & \textbf{79.7} & 73.5 & \textbf{74.1} & \textbf{74.1} \\
\bottomrule
\end{tabular}%
\end{table*}

\begin{table*}[t]
\centering
\caption{\textbf{Additional Language Benchmark Results.} ARC-Challenge (ARC-C) is evaluated with 25 shots using accuracy and normalized accuracy; CoQA and XQuAD-en are evaluated zero-shot using exact match (EM). All scores are percentages.}
\label{tab:supp_additional_language}

\begin{tabular}{l|cccccccccccc}
\toprule
\multirow{2}{*}{\textbf{Method}}
 & \multicolumn{3}{c}{\textbf{ARC-C$^\text{Acc.}$}} & \multicolumn{3}{c}{\textbf{ARC-C$^\text{Norm Acc.}$}} & \multicolumn{3}{c}{\textbf{CoQA$^\text{EM}$}} & \multicolumn{3}{c}{\textbf{XQuAD-en$^\text{EM}$}} \\
\cmidrule(lr){2-4}\cmidrule(lr){5-7}\cmidrule(lr){8-10}\cmidrule(lr){11-13}
 & \textbf{4} & \textbf{8} & \textbf{16} & \textbf{4} & \textbf{8} & \textbf{16} & \textbf{4} & \textbf{8} & \textbf{16} & \textbf{4} & \textbf{8} & \textbf{16} \\
\midrule
Ouro-1.4B & \textcolor{gray}{56.1} & \textcolor{gray}{53.8} & \textcolor{gray}{41.6} & \textcolor{gray}{60.8} & \textcolor{gray}{58.0} & \textcolor{gray}{44.4} & \textcolor{gray}{69.1} & \textcolor{gray}{66.5} & \textcolor{gray}{45.1} & \textcolor{gray}{45.0} & \textcolor{gray}{45.3} & \textcolor{gray}{28.7} \\
~+ SFT & 54.7 & 53.5 & 53.1 & 59.3 & 58.8 & 57.3 & \textbf{69.5} & \textbf{68.8} & 64.3 & 46.3 & 43.5 & 32.2 \\
~+ DS & 53.7 & 54.3 & 53.0 & 58.4 & 59.0 & 56.5 & 69.0 & 67.7 & 63.8 & \textbf{49.8} & 46.0 & 40.7 \\
\rowcolor[HTML]{DAE3F5}
~+ RLD & \textbf{56.4} & \textbf{56.8} & \textbf{57.0} & \textbf{59.6} & \textbf{60.6} & \textbf{60.4} & 68.8 & 68.5 & \textbf{68.3} & 47.0 & \textbf{46.7} & \textbf{46.7} \\
\midrule
Ouro-2.6B & \textcolor{gray}{61.5} & \textcolor{gray}{60.8} & \textcolor{gray}{58.5} & \textcolor{gray}{66.0} & \textcolor{gray}{65.5} & \textcolor{gray}{61.4} & \textcolor{gray}{71.4} & \textcolor{gray}{70.8} & \textcolor{gray}{44.1} & \textcolor{gray}{45.7} & \textcolor{gray}{44.0} & \textcolor{gray}{23.8} \\
~+ SFT & 60.2 & 59.3 & 58.5 & 64.9 & 64.0 & 62.8 & 70.1 & 70.0 & 64.7 & 38.5 & 34.1 & 22.9 \\
~+ DS & 60.3 & 58.9 & 58.4 & 64.5 & 63.7 & 62.9 & 70.5 & 69.4 & 65.0 & 45.6 & 38.2 & 26.0 \\
\rowcolor[HTML]{DAE3F5}
~+ RLD & \textbf{61.2} & \textbf{61.1} & \textbf{61.1} & \textbf{65.6} & \textbf{65.6} & \textbf{65.7} & \textbf{72.0} & \textbf{72.0} & \textbf{71.8} & \textbf{45.8} & \textbf{47.2} & \textbf{47.1} \\
\bottomrule
\end{tabular}%
\end{table*}

\begin{table*}[t]
\centering
\caption{\textbf{ChartQA Results by Question Split.} Zero-shot relaxed accuracy (\%) on the Human and Augmented splits, together with the overall score. Column labels specify the recurrence settings of LoopVL.}
\label{tab:supp_chartqa_splits}

\begin{tabular}{l|ccccccccc}
\toprule
\multirow{2}{*}{\textbf{Method}}
 & \multicolumn{3}{c}{\textbf{Human}} & \multicolumn{3}{c}{\textbf{Augmented}} & \multicolumn{3}{c}{\textbf{Overall}} \\
\cmidrule(lr){2-4}\cmidrule(lr){5-7}\cmidrule(lr){8-10}
 & \textbf{H2L3} & \textbf{H3L3} & \textbf{H4L3} & \textbf{H2L3} & \textbf{H3L3} & \textbf{H4L3} & \textbf{H2L3} & \textbf{H3L3} & \textbf{H4L3} \\
\midrule
LoopVL & \textcolor{gray}{67.8} & \textcolor{gray}{46.2} & \textcolor{gray}{11.6} & \textcolor{gray}{84.7} & \textcolor{gray}{67.5} & \textcolor{gray}{17.8} & \textcolor{gray}{76.3} & \textcolor{gray}{56.8} & \textcolor{gray}{14.7} \\
~+ SFT & 64.4 & 62.3 & 56.1 & \textbf{85.0} & 83.7 & 75.8 & \textbf{74.7} & 73.0 & 65.9 \\
~+ DS & 51.8 & 51.7 & 45.1 & 70.7 & 70.6 & 64.1 & 61.2 & 61.1 & 54.6 \\
\rowcolor[HTML]{DAE3F5}
~+ RLD & \textbf{65.1} & \textbf{63.9} & \textbf{62.2} & 83.8 & \textbf{84.1} & \textbf{83.0} & 74.4 & \textbf{74.0} & \textbf{72.6} \\
\bottomrule
\end{tabular}%
\end{table*}
We have supplemented text and vision results on MMLU-subset (Table \ref{tab:supp_mmlu_groups}), ARC-Challenge, CoQA, XQuAD (Table \ref{tab:supp_additional_language}), and ChartQA-subset (Table \ref{tab:supp_chartqa_splits}). On these benchmarks, RLD also maintains promising stability results. Specifically, on ARC-Challenge, RLD achieves the optimal results for both Ouro-1.4B and Ouro-2.6B across all the depth settings, showing good scalability for larger models. On the Augmented subset of ChartQA, RLD's performance drop under H4L3 is merely 2.0\%, while SFT and DS face 10.5\% and 24.3\%, respectively. These results further show the superiority of RLD on different models and benchmarks.

\section{Extended Visualizations}
\subsection{Latent Dynamics}
\label{sec:supp_vis_latent_dynamics}
\begin{figure*}[t]
    \centering
    \includegraphics[width=\linewidth]{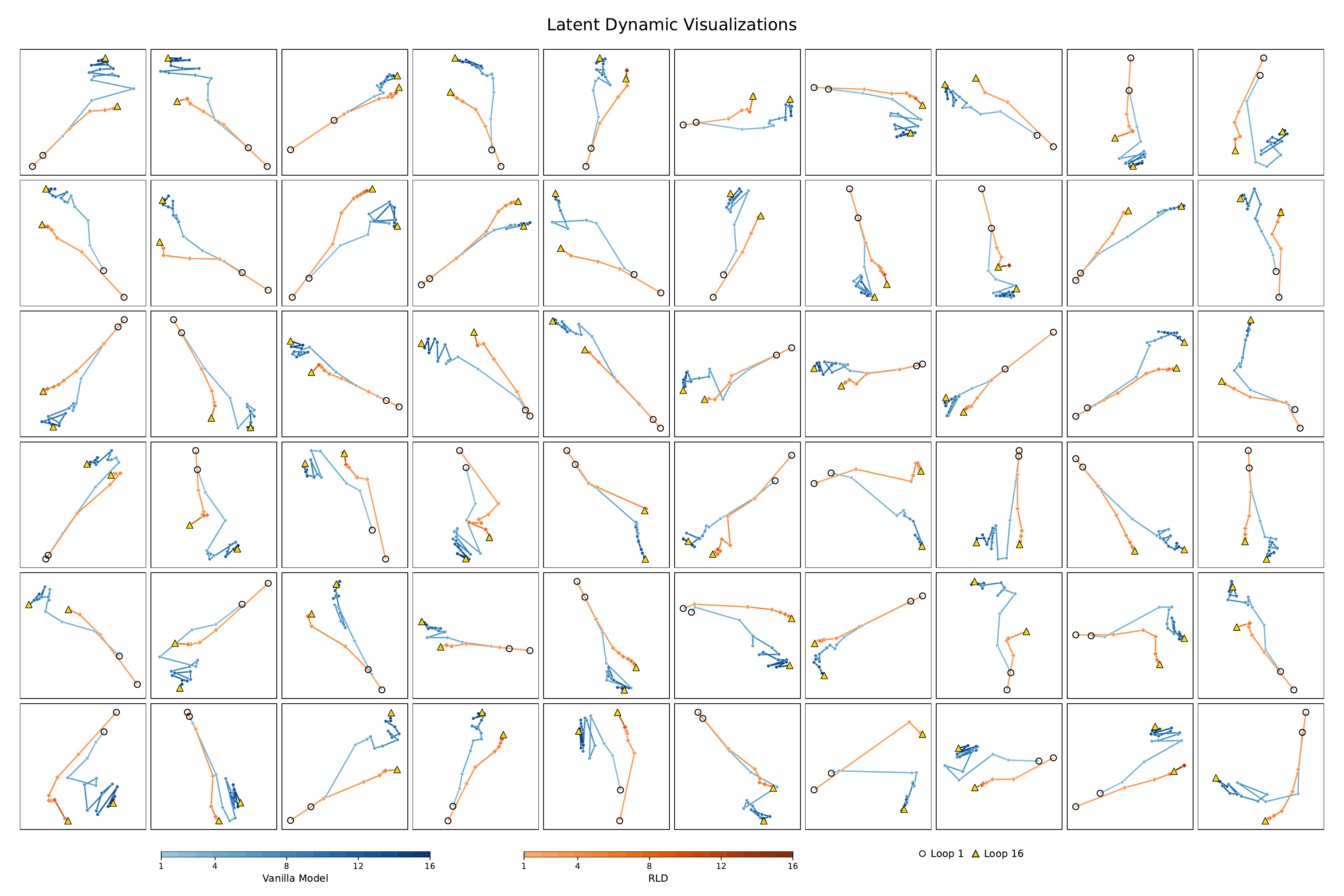}
    \caption{\textbf{Extended Visualizations on Latent Dynamics}. We adopt the same visualization setting as Fig. \ref{fig:exp_rectified_traces}.}
    \label{fig:supp_trace_vis}
\end{figure*}

We plot more token traces in Fig. \ref{fig:supp_trace_vis}. Sharing a similar trend with Fig. \ref{fig:exp_rectified_traces}, the vanilla model demonstrates a chaotic and random pattern, with most tokens vibrating intensely in the late phase. In contrast, the traces of RLD are consistent and smooth. Notably, different tokens have different steps before approaching the end, indicating that RLD assigns different thinking efforts for tokens with different difficulty, rather than simply constraining the late-stage update.

\subsection{Different Task Trajectories}
\label{sec:supp_vis_task_traj}
\begin{figure*}
    \centering
    \includegraphics[width=\linewidth]{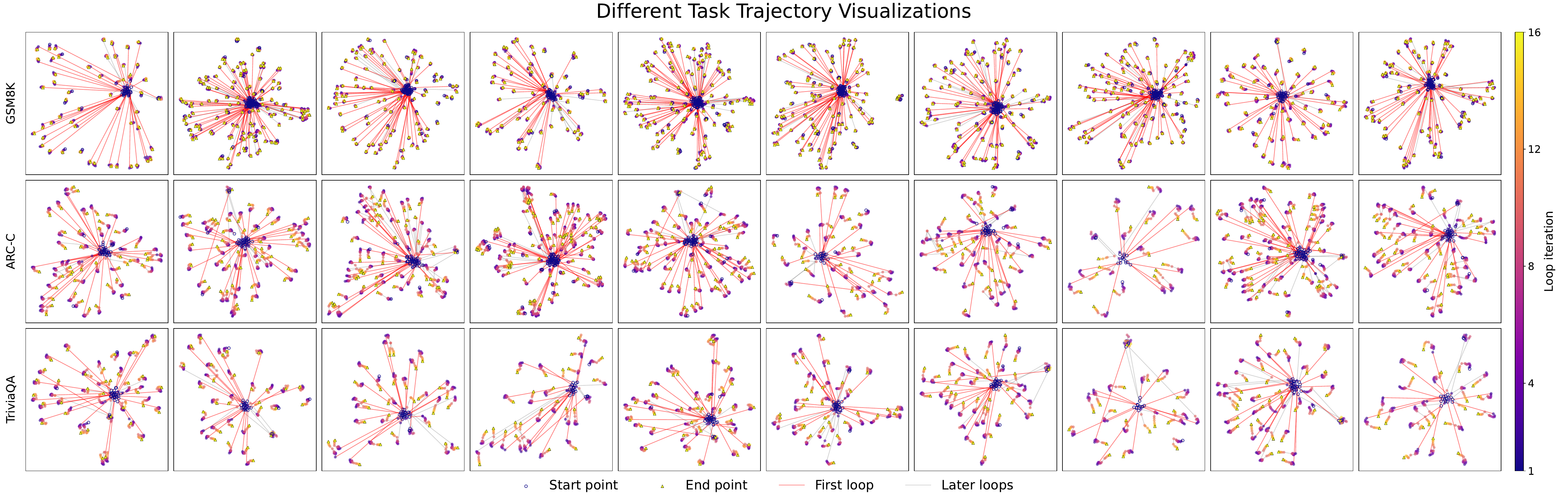}
    \caption{\textbf{Extended Visualizations on Different Task Trajectories.} We adopt the same visualization setting as Fig. \ref{fig:intro_tarj_vis}.}
    \label{fig:supp_trajectory_on_tasks}
\end{figure*}

As shown in Fig. \ref{fig:supp_trajectory_on_tasks}, we visualize the token traces of multiple tasks. In most cases, the first loop takes up the largest update, while the subsequent updates are small. This indicates that the model prefers to solve the problem within few loops, and more loops can make limited differences. Besides, this phenomenon happens on various tasks, constraining the model's step-by-step reasoning ability on hard problems.

\subsection{Attention Map in LoopVL}
\label{sec:supp_attention_map}
\begin{figure*}[t]
    \centering
    \begin{minipage}{\textwidth}
        \centering
        \includegraphics[width=\linewidth]{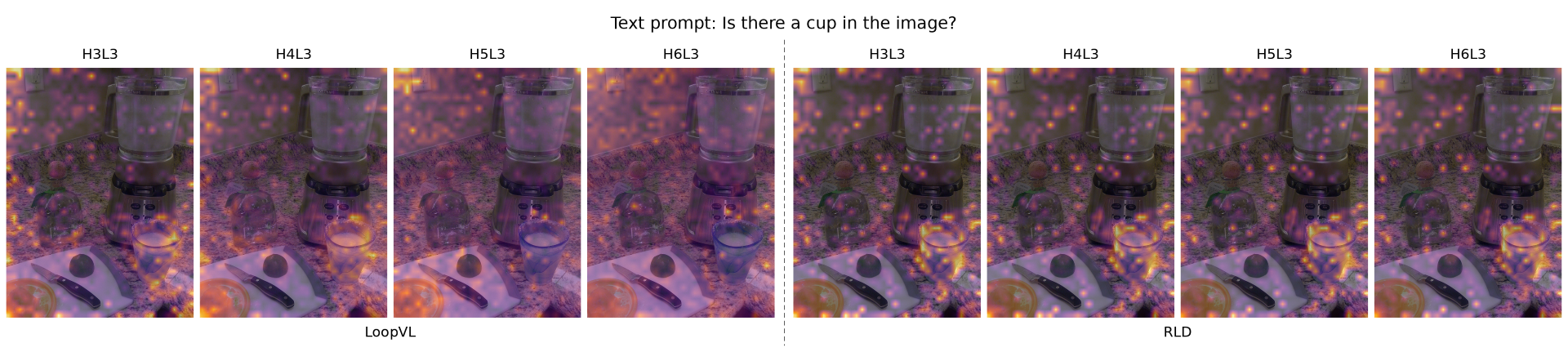}
    \end{minipage}
    \hfill
    \begin{minipage}{\textwidth}
        \centering
        \includegraphics[width=\linewidth]{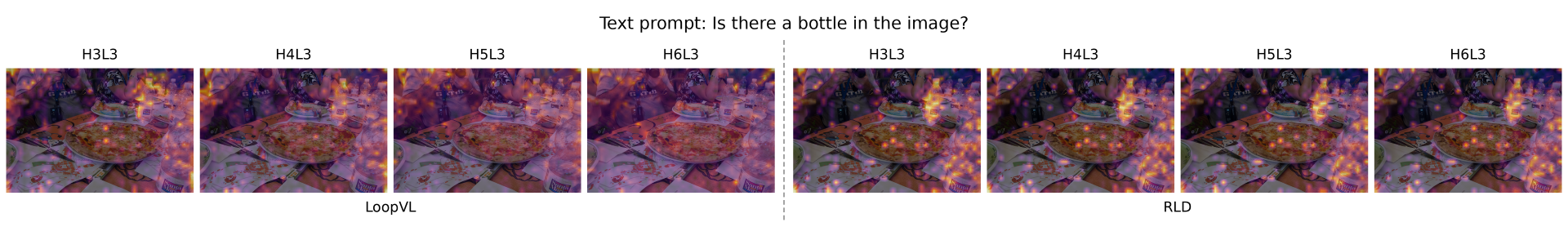}
    \end{minipage}
    \hfill
    \begin{minipage}{\textwidth}
        \centering
        \includegraphics[width=\linewidth]{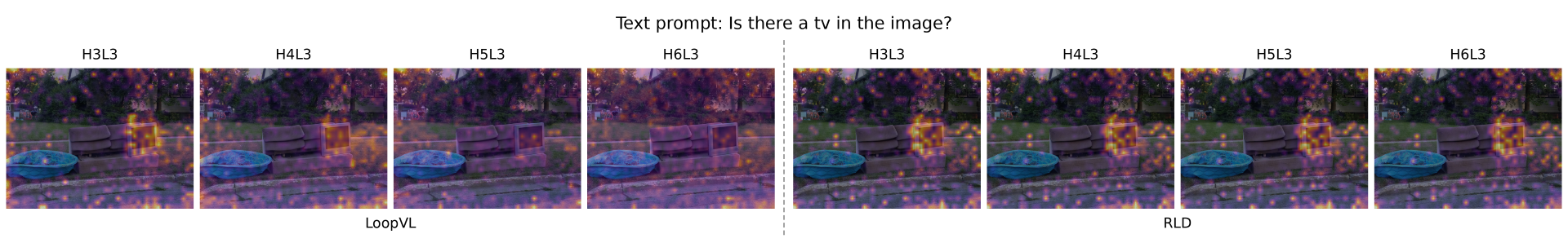}
    \end{minipage}
    \hfill
    \begin{minipage}{\textwidth}
        \centering
        \includegraphics[width=\linewidth]{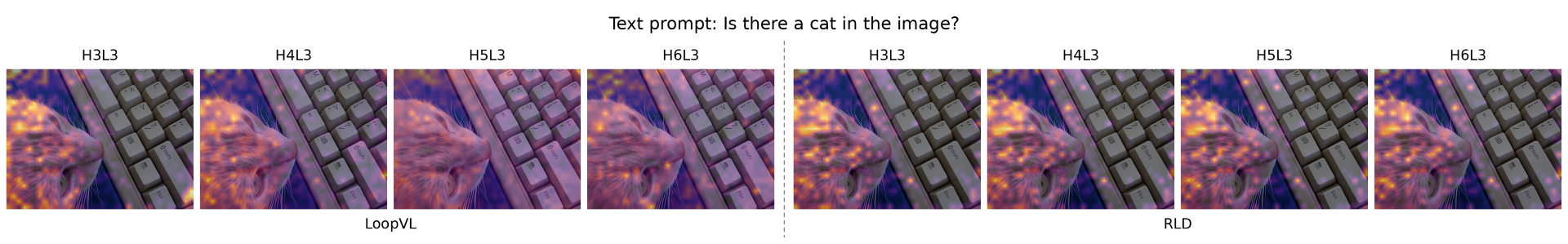}
    \end{minipage}
    \hfill
    \begin{minipage}{\textwidth}
        \centering
        \includegraphics[width=\linewidth]{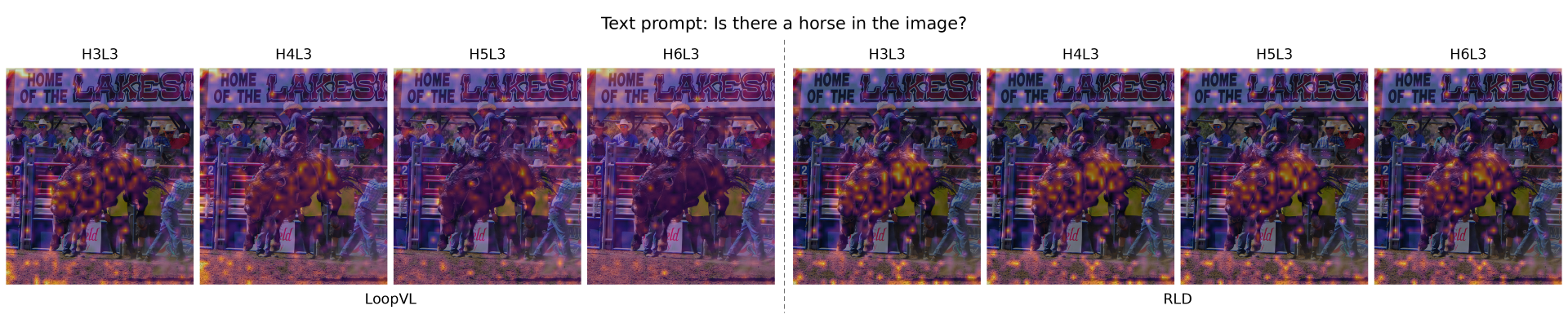}
    \end{minipage}
    \hfill
    \begin{minipage}{\textwidth}
        \centering
        \includegraphics[width=\linewidth]{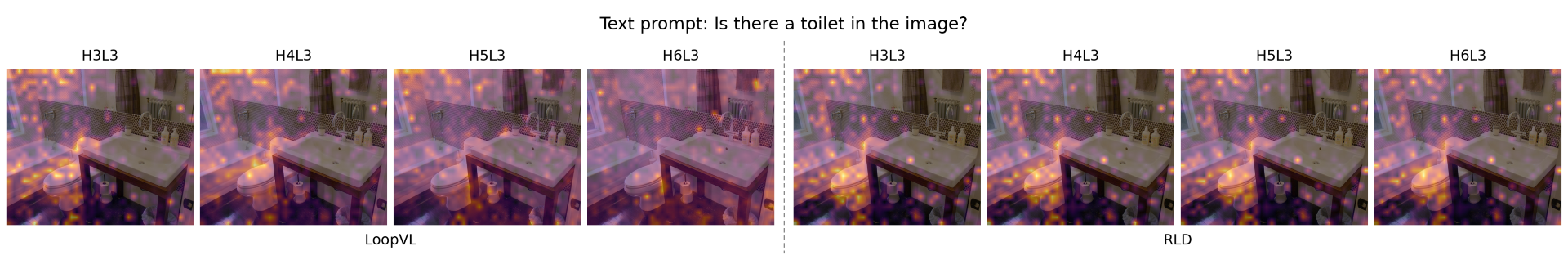}
    \end{minipage}
    \hfill
    \begin{minipage}{\textwidth}
        \centering
        \includegraphics[width=\linewidth]{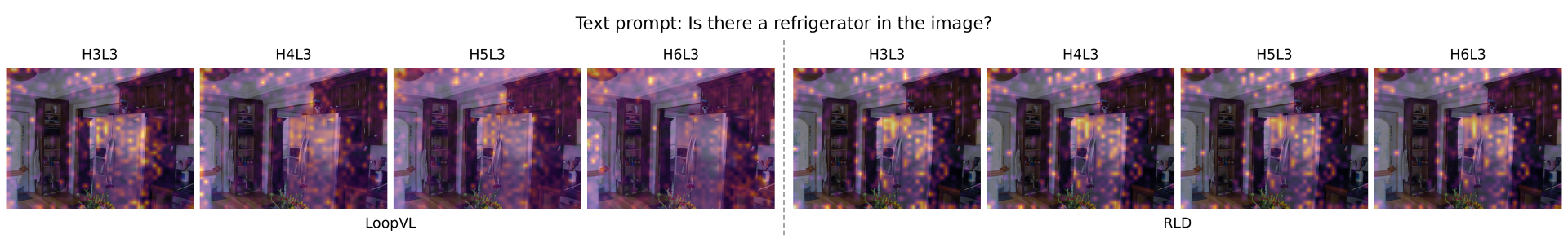}
    \end{minipage}
    \caption{\textbf{Extended Visualizations on Attention Map.}}
    \label{fig:supp_atten_map}
\end{figure*}

We randomly choose samples from the POPE \cite{li-etal-2023-evaluating} dataset and plot the attention map of the vanilla model and RLD in Fig. \ref{fig:supp_atten_map}. In these visualizations, the vanilla model is able to keep its attention on the key region under a lower loop. However, as the loop setting grows, it gradually loses its focus and attends to tokens unrelated to the question. In contrast, whether the loop setting is suitable or deeper than those seen during training, RLD consistently focuses on the regions most related to the question, and the distribution shift between loops is quite small. These qualitative results show the robustness of RLD under vision tasks.

\end{document}